\documentclass{article}

\usepackage{arxiv}

\usepackage[utf8]{inputenc} 
\usepackage[T1]{fontenc}    
\usepackage{hyperref}       
\usepackage{url}            
\usepackage{booktabs}       
\usepackage{amsfonts}       
\usepackage{nicefrac}       
\usepackage{microtype}      
\usepackage{lipsum}		
\usepackage{graphicx}
\usepackage{natbib}
\usepackage{doi}

\usepackage{amsthm}
\usepackage{algorithm}
\usepackage{algorithmic}
\newtheorem{theorem}{Theorem}
\newtheorem{lemma}[theorem]{Lemma}
\newtheorem{assumption}[theorem]{Assumption}
\newtheorem{corollary}[theorem]{Corollary}
\usepackage{multirow}
\usepackage{adjustbox}
\usepackage{framed}
\usepackage{amsmath}
\usepackage{wrapfig}
\usepackage[table]{xcolor}

\usepackage{rotating}
\usepackage{booktabs}
\newcommand{\jygl}{\cellcolor{gray!15}}

\title{Weight Concentration Regularization for Improving Pruning Robustness Under High Sparsity}

\author{
  Vincent-Daniel Yun\textsuperscript{1}, 
  Junhyuk Jo\textsuperscript{2}, 
  Sunwoo Lee\textsuperscript{2}\thanks{Corresponding Author. Under Review.} \\ \\
  \textsuperscript{1}University of Southern California, USA\\ 
  yunjuyou@usc.edu \\
  \textsuperscript{2}Inha University, Republic of Korea \\
  \{911whwnsgur, sunwool\}@inha.ac.kr  \\
}

\date{}

\renewcommand{\shorttitle}{}

\hypersetup{
pdftitle={Weight Concentration Regularization}
}

\begin{document}
\maketitle

\begin{abstract}
Deep neural networks achieve outstanding performance across vision and language tasks, yet their large parameter counts limit deployment in resource-constrained settings. One-shot pruning reduces model size without retraining, but models trained with standard objectives often suffer substantial accuracy drops under aggressive sparsity. Prior work mitigates this drop along two directions: regularizers such as $\ell_1$ and DeepHoyer that shape the weight distribution during training, and pruning-robust optimizers such as SAM, CrAM, and S$^2$SAM that flatten the loss landscape. However, existing regularizers either shrink all weights uniformly ($\ell_1$) or induce scale-invariant sparsity (DeepHoyer), without concentrating weight energy onto a small set of informative parameters. We propose a Weight Concentration Regularizer (WCR), a training-time regularizer that amplifies the magnitude of a small subset of parameters while driving the remainder toward zero, so that magnitude pruning predominantly removes parameters with negligible functional contribution. We provide a convergence analysis and evaluate WCR on LLM fine-tuning, image classification, and medical segmentation, demonstrating consistent improvements in pruning robustness across architectures and compatibility with existing pruning-robust optimizers.
\end{abstract}


\section{Introduction}
Deep neural networks have achieved strong performance across a wide range of domains, including computer vision, natural language processing, and speech recognition. Increasing model size improves representational capacity and performance; however, training and deployment require substantial computational resources, limiting applicability in resource-constrained environments. Model pruning aims to reduce model size by removing less important parameters while preserving performance~\cite{pruning}. Maintaining accuracy typically relies on iterative pruning with retraining~\cite{frankle2019lottery}, which is effective but computationally expensive. One-shot pruning offers a simpler alternative~\cite{one-shot}, removing redundant parameters in a single step without retraining or fine-tuning~\cite{he2017channel}. However, models trained with standard objectives often suffer significant performance degradation.

Prior work addresses this limitation from several perspectives. Regularization-based approaches, such as $\ell_1$ regularization and DeepHoyer~\cite{yang2020deephoyer}, encourage sparsity in the weight distribution, while constraint-based methods such as SFW~\cite{sfw} directly enforce sparsity during optimization. Pruning-robust optimizers, including SAM~\cite{sam}, CrAM~\cite{cram}, and S$^2$SAM~\cite{ssam}, improve generalization by promoting flat minima.  However, these approaches primarily focus on inducing sparsity or improving generalization, and do not explicitly address the trade-off between sparsity and information preservation. Standard sparsity-inducing methods tend to uniformly shrink weights, reducing overall weight energy and potentially weakening important parameters. As a result, when weight importance is diffusely distributed across parameters, removing small-magnitude weights can still lead to performance degradation.

To address this challenge, we seek to concentrate weight energy onto a small subset of parameters while driving the remaining weights toward zero. Under such a distribution, magnitude pruning can remove low-magnitude parameters with less performance degradation.

\begin{figure*}[t]
\centering 
\includegraphics[width=0.5\linewidth]{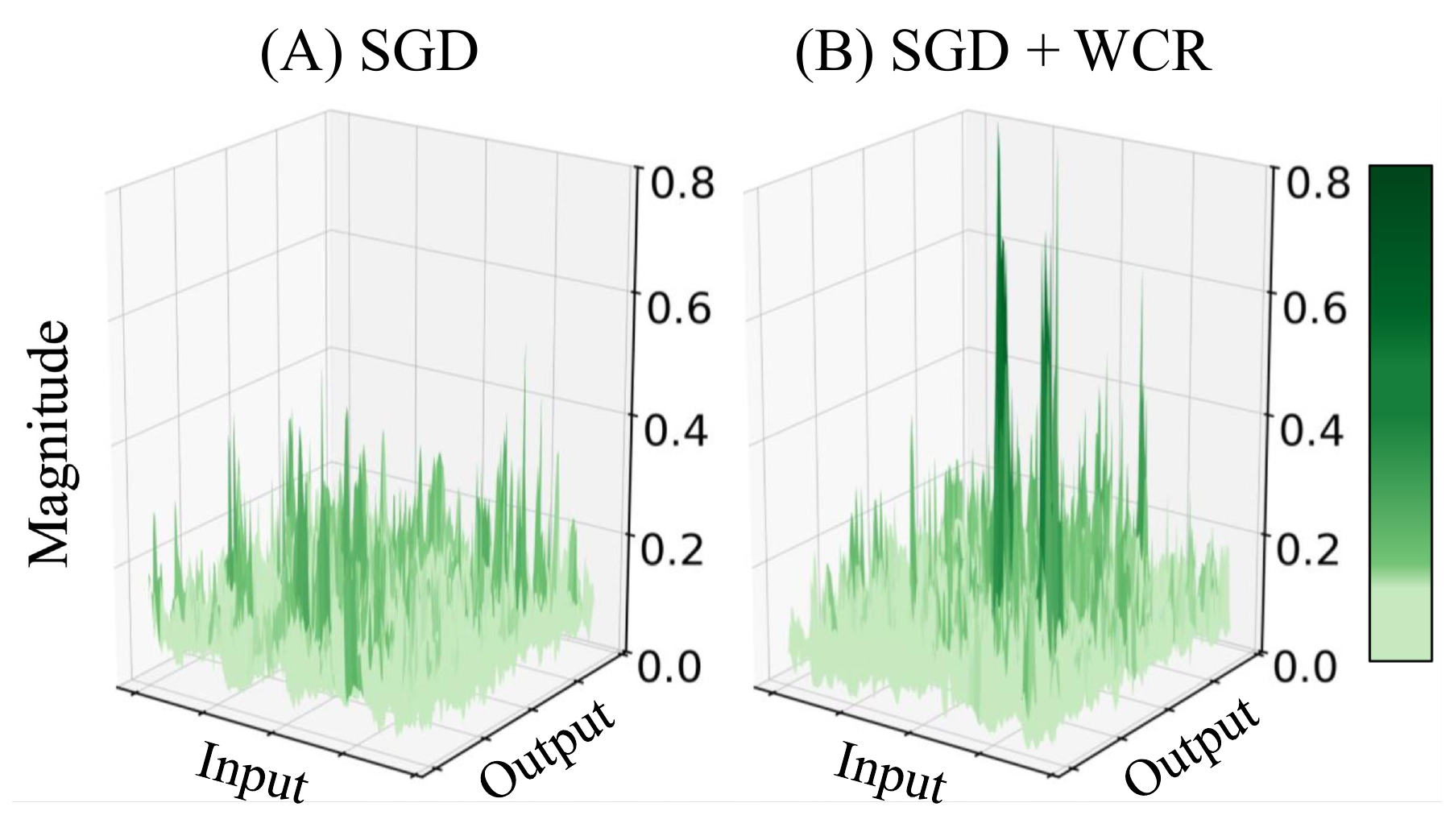} 
\caption{Visualization of parameter magnitudes of ResNet-18~\cite{resnet} on CIFAR-10~\cite{CIFAR} trained with and without the proposed Weight Concentration Regularizer (WCR). The horizontal axes correspond to input and output channel indices, and the vertical axis represents the magnitude of each parameter. WCR concentrates weight magnitude onto a small subset of parameters.} 

\label{fig:preliminary} 
\end{figure*}

In this work, we propose \textit{Weight Concentration Regularizer (WCR)}, a training-time regularizer that concentrates weight energy onto a small subset of parameters while driving the remaining weights toward zero. Unlike conventional sparsity-inducing regularizers, WCR explicitly encourages concentrated weight distributions rather than uniformly shrinking all parameters.

WCR achieves this by reshaping the weight magnitude distribution into a heavy-tailed form. Specifically, it penalizes the reciprocal of the variance of absolute weights, serving as a surrogate for weight concentration: enlarging this variance amplifies a small number of weights while suppressing the rest. WCR can be incorporated into standard stochastic gradient descent (SGD) and is compatible with existing pruning-robust optimizers. By concentrating weight energy during training, WCR enables magnitude pruning to remove parameters with minimal functional impact, improving robustness under aggressive pruning.

We evaluate WCR across multiple settings, including LLM fine-tuning, image classification, and medical image segmentation, demonstrating consistent improvements in pruning robustness across architectures and tasks. In addition, we provide a theoretical analysis showing that WCR preserves the standard convergence rate of SGD under mild smoothness assumptions.

\noindent
\textbf{Our contributions} are summarized as follows.
\begin{itemize}
    \item We introduce Weight Concentration Regularizer (WCR), a training-time regularizer that concentrates weight energy onto a small subset of parameters for improved one-shot pruning robustness.
    \item We empirically demonstrate that WCR improves pruning robustness across vision tasks, including image classification and medical image segmentation, and further show that it enhances existing pruning-robust optimizers while also improving robustness in LLM fine-tuning.
\end{itemize}

\footnotetext{Official code repository will be released soon.}

\section{Related works}

\noindent \textbf{Neural Network Pruning.}
Model pruning methods commonly drop a subset of model parameters to enable neural network training on resource-constrained systems~\cite{pruning}.
Early pruning approaches, such as Optimal Brain Damage~\cite{lecun1990optimal} and Optimal Brain Surgeon~\cite{hassibi1993second} pruned parameters based on the second-order sensitivity information to minimize the increase in loss.
Recently, several studies have proposed parameter magnitude-based pruning methods, typically using an $L_1$-norm criterion to remove parameters with small magnitudes.
Han et al.~\cite{han2015learning} demonstrated that such unstructured, magnitude-based methods can achieve high sparsity but often require fine-tuning.
In contrast, structured pruning methods remove entire channels or filters to reduce computational cost while preserving the key structural characteristics of model parameters~\cite{li2016pruning, he2017channel}.
More recent studies have explored dynamic and adaptive pruning approaches~\cite{Gao2024BilevelPruning}, which adjust sparsity during training to balance compactness and performance.

\noindent \textbf{One-Shot Pruning.}
One-shot pruning methods have been highlighted recently, which remove less important parameters in a single step after training~\cite{one-shot}.
A common direction is to shape the weight distribution during training through sparsity-inducing regularization. $L_1$ regularization shrinks all weights uniformly toward zero, while DeepHoyer~\cite{yang2020deephoyer} uses the scale-invariant ratio $\|w\|_1/\|w\|_2$ to protect large weights from shrinkage. Both target sparsity without concentrating weight energy on important parameters; in contrast, our regularizer concentrates weight energy on a small subset of weights while pushing the rest toward zero.
SFW~\cite{sfw} formulates pruning as a constrained optimization problem with a $K$-sparse constraint.
Recently, Sharpness-Aware Minimization (SAM)~\cite{sam} is known to not only improve generalization but also make models more robust to model pruning~\cite{sampruning}.
CrAM~\cite{cram} further enhances compression efficiency through perturbation-based optimization, and S$^2$SAM~\cite{ssam} achieves similar effects with a single-step approximation.
These methods improve pruning robustness mainly through loss-landscape smoothing and perturbation-based optimization. In contrast, our work focuses on reshaping the weight distribution during training, and can be combined with these approaches.
There have been several initialization-based pruning methods.
SNIP~\cite{snip} identifies important connections by measuring each weight's influence on the loss using a small batch of data, whereas GraSP~\cite{grasp} preserves gradient flow to maintain signal propagation during the early training stage.
SynFlow~\cite{SynFlow} selects critical weights by maximizing synthetic information flow based on absolute weight values.
Although these methods are effective for early-stage sparsification, they do not explicitly consider improving pruning robustness during training.
Therefore, we exclude these initialization-based methods from our comparison.

\noindent
\textbf{Weight Diversity and Structural Robustness.}
Maintaining weight diversity which can be represented as rank has been shown to improve model stability and generalization.
Soft Orthogonality (SO)~\cite{SO} and Double Soft Orthogonality (DSO)~\cite{DSO} preserve the effective rank of weights by constraining their Gram matrix toward the identity, but they rely on costly matrix-level computations that limit scalability. SRIP~\citep{DSO, candes2005decoding} minimizes the spectral norm of the difference between the Gram matrix of a weight matrix and the identity matrix.
While these methods improve model generalization, they do not explicitly encourage concentrated weight distributions for pruning. In contrast, our work concentrates weight magnitude onto a small subset of parameters during training, reshaping the weight distribution so that magnitude pruning preferentially removes low-magnitude weights.
\section{Method}
In this study, we introduce Weight Concentration Regularizer (WCR), a training-time regularizer that encourages weight magnitude to concentrate on a small subset of parameters while driving the rest toward zero. As a result, most of the weight energy becomes concentrated on a few high-magnitude parameters, while the remaining parameters become negligible. Under this distribution, magnitude pruning predominantly removes low-magnitude parameters while preserving the dominant weights.

\noindent\textbf{Preliminaries.} 
We denote the training dataset as $\mathcal{D} = \{(x_i, y_i)\}_{i=1}^M$, where each input $x_i \in \mathbb{R}^k$ and label $y_i$ belongs to the label space $\mathcal{Y}$. 
A neural network is parameterized by weights $w \in \mathbb{R}^d$, and its output for input $x$ is written as $f(x; w)$. The empirical loss is $L(w) = \frac{1}{M} \sum_{i=1}^M \ell(f(x_i; w), y_i)$, where $\ell(\cdot, \cdot)$ denotes a loss function. We let $\nabla L(w)$ denote the full gradient at $w$, and $\nabla L_t(w)$ the stochastic gradient computed from a minibatch at iteration $t$.

\noindent\textbf{Weight Concentration Regularizer (WCR).}
WCR concentrates weight energy by enlarging the variance of absolute weights during training. To target the main trainable weight tensors, WCR is applied to tensors of dimension greater than one (convolutional kernels and fully-connected weights) and excludes one-dimensional vectors such as biases and batch-norm parameters.
Let $\mathcal{W}=\{w^{(1)},\dots,w^{(L)}\}$ be the selected layer-wise weights. For each layer $w^{(\ell)}=(w^{(\ell)}_1,\dots,w^{(\ell)}_{n_\ell})$, we apply a smooth absolute-value transform for differentiability at the origin:
\begin{align}
    \tilde{w}^{(\ell)}_i = \sqrt{ (w^{(\ell)}_i)^2 + r } \approx |w^{(\ell)}_i|, \qquad r = 10^{-8}.
\end{align}
The variance is computed per-layer to handle scale differences across layers,
$\mathrm{Var}(\tilde{w}^{(\ell)}) = \frac{1}{n_\ell} \sum_{i=1}^{n_\ell}
(\tilde{w}^{(\ell)}_i - \frac{1}{n_\ell}\sum_{j=1}^{n_\ell} \tilde{w}^{(\ell)}_j)^2$, 
and the penalty is the sum of layer-wise reciprocal variances:
\begin{equation}
\psi(w) = 
\sum_{w^{(\ell)} \in \mathcal{W}} 
    \frac{1}{\mathrm{Var}(\tilde{w}^{(\ell)}) + \epsilon},
\end{equation}
with $\epsilon = 10^{-8}$ for numerical stability. Since $\psi(w)$ depends only on weights, $\nabla \psi(w)$ is deterministic and does not contribute to stochastic gradient variance. The total training objective is
\begin{equation}
L_{total}(w) = L(w) + \lambda \psi(w), \label{eq:var}
\end{equation}
where $\lambda > 0$ is a balancing coefficient. Optimization follows the standard SGD update $w_{t+1} = w_t - \eta (\nabla L_{t}(w_t) + \lambda \nabla\psi(w_t))$, with $\eta > 0$ the learning rate.

\begin{figure*}[t]
  \centering
  \includegraphics[width=1.0\linewidth]{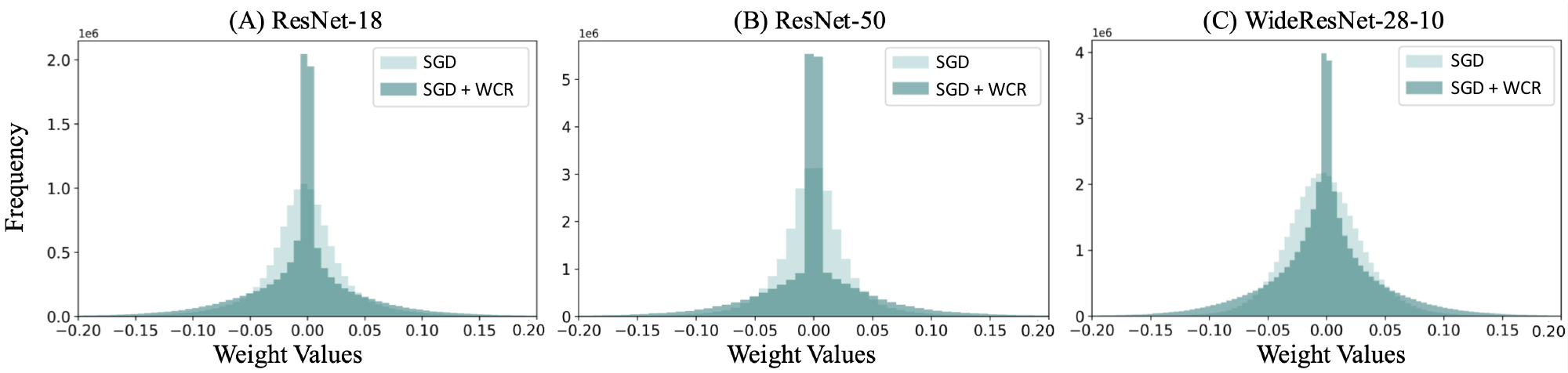}
  \caption{Weight-parameter distributions for models trained with standard SGD and SGD with WCR ($\lambda=10^{-5}$). (A) ResNet-18/CIFAR-10, (B) ResNet-50/SVHN, (C) WideResNet-28-10/CIFAR-100. WCR yields a heavy-tailed distribution: a sharper peak near zero alongside a wider tail of large-magnitude weights, indicating simultaneous concentration and sparsification.}
  \label{fig:weightdist}
\end{figure*}

\noindent\textbf{How WCR concentrates weight energy.}
The penalty $\psi(w)$ admits a closed-form expression that provides insight into how WCR promotes both sparsity and weight concentration. For each layer, the variance of absolute weights can be written as
\begin{align}
\mathrm{Var}(\tilde{w}^{(\ell)})
&= \frac{\|\tilde{w}^{(\ell)}\|_2^2}{n_\ell} - \frac{\|\tilde{w}^{(\ell)}\|_1^2}{n_\ell^2}
= \frac{\|\tilde{w}^{(\ell)}\|_2^2 \cdot \big(n_\ell - H(\tilde{w}^{(\ell)})\big)}{n_\ell^2},
\end{align}
where $H(u) := \|u\|_1^2/\|u\|_2^2 \in [1, n_\ell]$ is the Hoyer sparsity measure.

The corresponding penalty becomes
\begin{equation}
\frac{1}{\mathrm{Var}(\tilde{w}^{(\ell)}) + \epsilon}
= \frac{n_\ell^2}{\|\tilde{w}^{(\ell)}\|_2^2 \cdot \big(n_\ell - H(\tilde{w}^{(\ell)})\big) + n_\ell^2 \epsilon}.
\end{equation}

This expression highlights that the penalty depends jointly on the total $L_2$ energy $\|\tilde{w}^{(\ell)}\|_2^2$ and the sparsity gap $n_\ell - H(\tilde{w}^{(\ell)})$. Increasing either term enlarges the denominator, thereby reducing the penalty. In practice, this encourages a distribution in which a small subset of parameters carries most of the weight energy while the remaining parameters become small.

Unlike scale-invariant sparsity regularizers such as DeepHoyer~\cite{yang2020deephoyer}, which primarily act on the sparsity structure, WCR also depends on the overall energy scale. Empirically, this leads to weight distributions where a small number of parameters dominate the total energy.
\section{Theoretical analysis}
This section analyzes the convergence behavior of stochastic gradient descent (SGD) when optimizing the proposed Weight Concentration Regularizer (WCR) objective. By extending the classic SGD convergence analysis~\cite{sgdproof}, we show that the total objective \(L_{\mathrm{total}}(w) = L(w) + \lambda \psi(w)\) satisfies the standard convergence-to-stationarity guarantees under the $\beta_{1}$- and $\beta_{2}$-smoothness assumptions and standard bounded-variance conditions. Detailed proofs are provided in the Appendix~\ref{appendix-theory}.

\begin{assumption}[$\beta_1$-smoothness]
\label{assumption:smoothness}
The loss function \(L:\mathbb{R}^d \to \mathbb{R}\) is \(\beta_1\)-smooth, i.e., its gradient is \(\beta_1\)-Lipschitz continuous $\|\nabla L(u) - \nabla L(v)\| \leq \beta_1 \|u-v\|, \forall u,v \in \mathbb{R}^d.$
Equivalently, for any \(u,v \in \mathbb{R}^d\), the following descent lemma holds:
\begin{align}
L(u) \leq L(v) + \langle \nabla L(v), u-v \rangle + \tfrac{\beta_1}{2}\|u-v\|^2.
\end{align}
\end{assumption}

\begin{assumption}[Unbiased stochastic gradient]
\label{assumption:unbiased}
At each iteration \(t\), the stochastic gradient \(\nabla L_t(w_t)\) is an unbiased estimator of the true gradient:
\begin{align}
\mathbb{E}[\nabla L_t(w_t)] = \nabla L(w_t).
\end{align}
\end{assumption}


\begin{assumption}[Local smoothness of $\psi(w)$]
\label{assumption:Bsmoothness}
The regularizer $\psi:\mathbb{R}^d \to \mathbb{R}$ is continuously differentiable. 
Moreover, we assume that $\psi(w)$ has a $\beta_2$-Lipschitz continuous gradient over a bounded domain $\mathcal{W}$ containing the iterates $\{w_t\}$, i.e.,
\begin{align}
\|\nabla \psi(w_1) - \nabla \psi(w_2)\| \le \beta_2 \|w_1 - w_2\|, \quad \forall w_1, w_2 \in \mathcal{W}.
\end{align}
\end{assumption}


\begin{lemma}
Under the $\beta_{1}$-smoothness of $L$ and the $\beta_{2}$-smoothness of $\psi$, 
the combined objective $L_{\mathrm{total}}(w)=L(w)+\lambda\psi(w)$
is $\beta$-smooth over $\mathcal{W}$ for $\beta \le \beta_{1}+\lambda\beta_{2}$.
Together with the bounded-variance conditions 
$\mathbb{E}[\nabla L_t(w_t)] = \nabla L(w_t)$ and 
$\mathbb{E}\!\left[\|\nabla L_t(w_t) - \nabla L(w_t)\|^2\right] \le \frac{\sigma^2}{b}$ (where the regularizer $\psi$ is deterministic, implying $\sigma_{\psi} = 0$),
if the step size satisfies $\eta \le 1/\beta$, then the classic SGD update
$w_{t+1} = w_t - \eta\big(\nabla L_t(w_t) + \lambda\nabla\psi(w_t)\big)$
guarantees:
\label{lemma1}
\begin{align}
\mathbb{E}\big[ L&(w_{t+1}) + \lambda \psi(w_{t+1}) \big] \le \mathbb{E}\big[ L(w_t) + \lambda \psi(w_t) \big]  - \frac{\eta}{2}\, \mathbb{E}\!\left[\|\nabla L(w_t) + \lambda \nabla\psi(w_t)\|^2\right] + \frac{\eta^2 \beta}{2b}\sigma^2.
\end{align}
\end{lemma}

\begin{theorem}
Under the same smoothness and bounded-variance conditions stated in Lemma~\ref{lemma1}, and for a fixed step size $0 < \eta \le 1/\beta$, the SGD update satisfies:
\begin{align}
\frac{1}{T} \sum_{t=0}^{T-1} &\mathbb{E}[\|\nabla L(w_t) + \lambda \nabla \psi(w_t)\|^2]  \\&\leq \frac{2}{T \eta} \big( (L(w_0) + \lambda \psi(w_0))\big)  - \frac{2}{T \eta} \big(\mathbb{E}[(L(w_T) + \lambda \psi(w_T))] \big) + \frac{\eta \beta \sigma^2}{2b}
\end{align}
\end{theorem}

\begin{corollary}
Under the same smoothness and bounded-variance assumptions stated in 
Lemma~\ref{lemma1}, and assuming further that $L_{\mathrm{total}}$ is bounded below by $L^{\inf}_{\mathrm{total}} > -\infty$, let the step size be constant $\eta = \tfrac{c}{\sqrt{T}}$ with $0 < c \le \tfrac{1}{\beta}$. Then the SGD iterates satisfy:
\begin{align}
\frac{1}{T} \sum_{t=0}^{T-1}
&\mathbb{E}[\big\| \nabla L(w_t) + \lambda \nabla \psi(w_t)\big\|^2] \le \frac{2\big((L(w_0)+\lambda\psi(w_0))-L^{\inf}_{total}\big)}{c}\,T^{-1/2} \nonumber + \frac{\beta c}{2b}\sigma^2\,T^{-1/2}.
\end{align}
\end{corollary}


\noindent \textbf{Remark 1.}  
Under Assumption~\ref{assumption:Bsmoothness} on the differentiable concentration regularizer, the proposed objective function in (\ref{eq:var}) preserves the convergence guarantee of standard SGD. Since the regularizer is a deterministic function of the weights, it does not introduce additional stochastic gradient variance. Consequently, the overall convergence rate remains $\mathcal{O}(1/\sqrt{T})$, matching the rate reported in previous works~\cite{ghadimi2013stochastic,yu2019linear}. Therefore, the asymptotic convergence rate remains consistent with standard SGD.


\noindent\textbf{Remark 2.}
The smoothed transformation of the absolute value 
$\tilde{w}_i = \sqrt{w_i^2 + r}$ with $r>0$ is differentiable for all weight values.  
Since the layer-wise variance $\operatorname{Var}(\tilde{w}^{(\ell)})$ is a differentiable function of the corresponding layer parameters $w^{(\ell)}$, each reciprocal term 
$1/(\operatorname{Var}(\tilde{w}^{(\ell)}) + \epsilon)$ is also differentiable.  
Because the regularizer $\psi(w)$ is defined as the sum of these layer-wise terms, it is differentiable with respect to the full parameter vector $w$.  
In our analysis, we further assume that $\psi$ satisfies the $\beta_2$-smoothness condition stated in Assumption~\ref{assumption:Bsmoothness}.

\noindent \textbf{Remark 3.} In practice, WCR uses a relatively small regularization coefficient $\lambda$, limiting the contribution of the additional regularization term.

\begin{corollary}[Diminishing step size]
\label{cor:diminishing}
Under the same smoothness and bounded-variance assumptions stated in Lemma~\ref{lemma1},
and assuming further that the total objective $L(w)+\lambda\psi(w)$ is bounded below by 
$L^{\inf}_{\mathrm{total}} > -\infty$, consider SGD with a diminishing step-size sequence $(\eta_t)_{t\ge0}$ satisfying $0<\eta_t \le \frac{1}{\beta}$, $\sum_{t=0}^\infty \eta_t = \infty$, and $\sum_{t=0}^\infty \eta_t^2 < \infty$
Then the (step-size)–weighted average of squared gradients vanishes when $\nabla L_\mathrm{{total}}(w_t)=\nabla L(w_t)+\lambda\nabla \psi(w_t)$ and $T$ is the total number of SGD iterations:
\begin{align}
\lim_{T\to\infty}\frac{1}{\sum_{t=0}^{T-1}\eta_t}
\sum_{t=0}^{T-1}\eta_t\,\mathbb{E}\left[\|\nabla L_{\mathrm{total}}(w_t)\|^2\right] = 0 \Rightarrow \liminf_{T\to\infty}\ \min_{0\le t\le T-1}\ \mathbb{E}\left[\|\nabla L_{\mathrm{total}}(w_t)\|^2\right]=0
\end{align}

Therefore, there exists a subsequence along which the expected squared gradient converges to $0$.
\end{corollary}

\noindent \textbf{Remark 4.}  
Corollary~\ref{cor:diminishing} indicates that, as the learning rate approaches zero, SGD drives the model parameters toward a stationary point of the objective. This behavior is consistent with that of standard SGD, showing that the asymptotic convergence behavior remains consistent with standard SGD under the stated assumptions.

The above analysis establishes that the proposed weight concentration regularizer does not affect the stability or convergence behavior of standard stochastic gradient descent. Under common smoothness and bounded-variance assumptions, the total objective \(L_{\mathrm{total}}(w) = L(w) + \lambda \psi(w)\) converges to a stationary point with the same \(\mathcal{O}(1/\sqrt{T})\) rate as standard SGD. 

\section{Experimental results}
\subsection{Experimental setup}

We design our experiments to evaluate two roles of WCR. First, we test whether WCR improves pruning robustness and composes with pruning-robust optimizers in vision tasks. This setting evaluates whether weight-distribution shaping provides gains beyond loss-landscape smoothing methods such as SAM~\cite{sam}, CrAM~\cite{cram}, and S$^2$SAM~\cite{ssam}. Second, we evaluate WCR as a standalone weight-shaping regularizer in LLM fine-tuning. Since AdamW + LoRA is the standard optimization pipeline for parameter-efficient fine-tuning, we compare WCR against $\ell_1$ and DeepHoyer~\cite{yang2020deephoyer} under the same optimizer and pruning protocol, isolating the effect of the regularization mechanism itself.

\noindent\textbf{Datasets and models.}
For image classification, we train ResNet-18~\cite{resnet} on CIFAR-10~\cite{CIFAR}, ResNet-50 on SVHN~\cite{SVHN}, WideResNet-28-10~\cite{wideres2} on CIFAR-100, and ImageNet-pretrained ViT-B/32~\cite{VIT} on CIFAR-100 and Tiny-ImageNet~\cite{tinyimagenet}. For segmentation, we train ResNet-50--UNet~\cite{unet} on the LGG Brain MRI dataset~\cite{lgg}. For LLM fine-tuning, we fine-tune Qwen-2.5-1.5B~\cite{qwen} and LLaMA-3.2-1B~\cite{llama3} on Commonsense-170K following the train/test protocol of DoRA~\cite{dora}. The training set is the union of eight commonsense reasoning datasets: BoolQ, PIQA, SIQA, HellaSwag, WinoGrande, ARC-Easy, ARC-Challenge, and OBQA. We evaluate on the corresponding test splits.

\noindent\textbf{Training setup.}
All vision models are trained for 200 epochs with batch size 128 for classification and 64 for segmentation. We set $\lambda{=}10^{-5}$ for CNNs and $\lambda{=}10^{-2}$ for ViT-B/32. Classification follows the original SFW/CrAM/SAM/S$^2$SAM setups. For segmentation and ViT-B/32, we use an initial learning rate of $0.001$ with $0.5\times$ step decay every 50 epochs, with early stopping for segmentation.

For LLMs, we fine-tune with LoRA using rank $r{=}32$, $\alpha{=}64$, and dropout $0.05$, applied to all attention and MLP projections (\texttt{q\_proj}, \texttt{k\_proj}, \texttt{v\_proj}, \texttt{o\_proj}, \texttt{gate\_proj}, \texttt{up\_proj}, \texttt{down\_proj}). Training uses AdamW with $\beta_1{=}0.9$, $\beta_2{=}0.999$, weight decay $0$, batch size 16, gradient accumulation 4, and learning rate $2{\times}10^{-4}$ with a linear schedule and 100 warmup steps. We train for 2 epochs with cutoff length 256. For both Qwen and LLaMA models, $\lambda{=}10^{-7}$ is used for $\ell_1$, while $\lambda{=}10^{-6}$ is used for DeepHoyer and WCR. Full $\lambda$ tuning results are provided in Appendix~\ref{llm-hyper}.

\noindent\textbf{One-shot pruning.}
All pruning is performed after training, with no retraining or fine-tuning. CNNs use global $\ell_1$ magnitude pruning over convolution weights. ViT-B/32 is pruned uniformly across Q, K, V, projection, and FFN layers. For LLMs, we apply $\ell_1$ magnitude pruning at $\{0\%,20\%,30\%\}$ over the LoRA-adapted attention and MLP projections.

\noindent\textbf{Evaluation metrics.}
For image classification, we report validation accuracy. For segmentation, we report F1, Tversky index, and Hausdorff distance. For LLM fine-tuning, we report test accuracy on each commonsense reasoning dataset and their average.

\begin{table*}[h]
\centering
\scriptsize
\caption{Accuracy (\%) for CNNs (CIFAR-10/100, SVHN) and ViT-B/32 (CIFAR-100, Tiny-ImageNet), without ('w/o') and with ('w') WCR. Results are averaged over 3 random seeds. SFW is not applicable to ViT-B/32.}
\vspace{3pt}
\begin{tabular}{llc>{\columncolor{gray!15}}cc>{\columncolor{gray!15}}cc>{\columncolor{gray!15}}cc>{\columncolor{gray!15}}c}
\toprule
& &
\multicolumn{2}{c}{\textbf{Dense}} &
\multicolumn{2}{c}{\textbf{92\% Pruned}} &
\multicolumn{2}{c}{\textbf{94\% Pruned}} &
\multicolumn{2}{c}{\textbf{96\% Pruned}} \\
\cmidrule(lr){3-4} \cmidrule(lr){5-6} \cmidrule(lr){7-8} \cmidrule(lr){9-10}
\textbf{Setting} & \textbf{Method} &
\textbf{w/o} & \textbf{w} &
\textbf{w/o} & \textbf{w} &
\textbf{w/o} & \textbf{w} &
\textbf{w/o} & \textbf{w} \\
\midrule
\multirow{5}{*}{\shortstack[l]{ResNet-18\\CIFAR-10}}
 & SGD     & 92.75 & 92.63 & 31.60 & 80.24 & 26.10 & 59.91 & 19.69 & 25.70 \\
 & SFW     & 92.54 & 92.34 & 92.54 & 92.32 & 92.22 & 92.29 & 84.15 & 89.93 \\
 & CrAM    & 94.19 & 93.86 & 86.54 & 92.14 & 80.40 & 89.47 & 66.66 & 81.72 \\
 & SAM     & 95.19 & 94.08 & 87.20 & 93.91 & 76.29 & 93.77 & 48.36 & 92.08 \\
 & S$^2$SAM & 95.23 & 93.71 & 91.02 & 93.67 & 87.21 & 93.59 & 72.51 & 93.44 \\
\midrule
\multirow{5}{*}{\shortstack[l]{ResNet-50\\SVHN}}
 & SGD     & 94.44 & 94.51 & 9.16  & 12.32 & 9.15  & 10.84 & 9.15  & 10.05 \\
 & SFW     & 95.79 & 95.80 & 95.82 & 95.79 & 95.73 & 95.77 & 18.20 & 95.49 \\
 & CrAM    & 95.97 & 95.79 & 78.51 & 90.51 & 53.30 & 82.65 & 27.78 & 31.26 \\
 & SAM     & 96.96 & 95.98 & 73.37 & 95.98 & 24.98 & 95.98 & 9.74  & 95.99 \\
 & S$^2$SAM & 96.88 & 95.78 & 73.92 & 95.78 & 39.55 & 95.78 & 17.28 & 95.78 \\
\midrule
\multirow{5}{*}{\shortstack[l]{WideRes-28-10\\CIFAR-100}}
 & SGD     & 75.93 & 75.82 & 2.86  & 10.88 & 1.00  & 4.00  & 1.00  & 1.00  \\
 & SFW     & 63.01 & 64.91 & 63.0  & 64.90 & 59.59 & 62.18 & 1.00  & 18.77 \\
 & CrAM    & 76.29 & 75.86 & 25.87 & 55.47 & 7.43  & 31.56 & 3.06  & 7.73  \\
 & SAM     & 78.43 & 79.51 & 4.26  & 77.95 & 1.94  & 73.97 & 1.00  & 56.56 \\
 & S$^2$SAM & 79.30 & 79.30 & 6.98  & 77.33 & 3.96  & 73.95 & 1.13  & 58.19 \\
\bottomrule
\toprule
& &
\multicolumn{2}{c}{\textbf{Dense}} &
\multicolumn{2}{c}{\textbf{60\% Pruned}} &
\multicolumn{2}{c}{\textbf{70\% Pruned}} &
\multicolumn{2}{c}{\textbf{80\% Pruned}} \\
\cmidrule(lr){3-4} \cmidrule(lr){5-6} \cmidrule(lr){7-8} \cmidrule(lr){9-10}
\textbf{Dataset} & \textbf{Method} &
\textbf{w/o} & \textbf{w} &
\textbf{w/o} & \textbf{w} &
\textbf{w/o} & \textbf{w} &
\textbf{w/o} & \textbf{w} \\
\midrule

 & SGD     & 89.84 & 89.75 & 86.30 & 88.58 & 78.33 & 85.66 & 51.13 & 73.69 \\
ViT-B/32 & CrAM    & 88.71 & 87.82 & 86.33 & 87.08 & 82.80 & 85.47 & 70.31 & 79.29 \\
CIFAR-100 & SAM     & 87.36 & 87.82 & 84.83 & 87.82 & 80.29 & 87.97 & 64.11 & 84.95 \\
 & S$^2$SAM & 88.06 & 87.90 & 84.82 & 86.70 & 81.95 & 86.80 & 69.58 & 85.25 \\
\midrule

 & SGD     & 86.89 & 86.72 & 82.90 & 85.43 & 73.79 & 79.23 & 39.59 & 60.15 \\
ViT-B/32 & CrAM    & 87.08 & 87.12 & 83.29 & 84.46 & 75.36 & 77.14 & 47.25 & 57.12 \\
Tiny-ImageNet & SAM     & 87.29 & 87.11 & 82.79 & 85.77 & 73.22 & 79.82 & 41.57 & 59.40 \\
 & S$^2$SAM & 87.20 & 87.31 & 82.62 & 84.59 & 72.86 & 79.90 & 41.12 & 59.37 \\
\bottomrule
\end{tabular}
\label{table:exp_cnn}
\end{table*}

\begin{table*}[t]
\scriptsize
\centering
\caption{Segmentation on LGG MRI (ResNet-50--UNet) at dense, 85\%, and 88\% pruning. Results are averaged over 3 random seeds. "--" indicates SGD failed to produce segmentation. }
\vspace{3pt}
\begin{adjustbox}{width=\textwidth}
\begin{tabular}{llccrlccrlccr} 
\toprule
&\textbf{Pruning Ratio}&  \multicolumn{3}{c}{\textbf{Dense}} 
&& \multicolumn{3}{c}{\textbf{85\% Pruned}} 
&& \multicolumn{3}{c}{\textbf{88\% Pruned}} 
\\
\cmidrule{3-5} \cmidrule{7-9} \cmidrule{11-13}
\textbf{Model} & \textbf{Method} & 
\textbf{F1} & \textbf{Tversky} & \textbf{H-Dist.}
&& 
\textbf{F1} & \textbf{Tversky} & \textbf{H-Dist.}
&& 
\textbf{F1} & \textbf{Tversky} & \textbf{H-Dist.}\\
\midrule
\multirow{10}{*}{Res50-Unet} & SGD          & 0.9284 & 0.9439 & 12.55 && 0.3083 & 0.3848 & 108.78 && - & - & -\\
& \jygl SGD + WCR            &\jygl 0.9258 &\jygl 0.9377 &\jygl 13.14 &\jygl&\jygl0.5136 &\jygl 0.5294 &\jygl 97.41 &\jygl&\jygl 0.5280 &\jygl 0.5253 &\jygl 87.55\\
                        & SFW                   & {0.9295} & 0.9452 & 9.18 && 0.5017 & 0.4480 & 89.97  && 0.1435 & 0.1100 & 84.42\\
                        & \jygl SFW + WCR          &\jygl 0.9191 &\jygl 0.9346 &\jygl 13.84 &\jygl&\jygl {0.7938} &\jygl 0.7928 &\jygl {32.61} &\jygl&\jygl {0.7210}  &\jygl 0.6909 &\jygl {31.78}\\
                        & CrAM                  & 0.9163 & 0.9366 & 19.11 && 0.1877 & 0.2758 & 110.31 && 0.2811 & 0.2888 & 105.89\\
                         & \jygl CrAM + WCR         &\jygl 0.8994 &\jygl 0.9245 &\jygl 21.12 &\jygl&\jygl 0.4066 &\jygl 0.5005 &\jygl 106.17 &\jygl&\jygl 0.3264 &\jygl 0.3345 &\jygl 103.99\\
                        & SAM                   & 0.9264 & 0.9378 & 9.26 && 0.3903 & 0.4399 & 107.11 && 0.2786 & 0.2587 & 100.13\\
                         & \jygl SAM + WCR          &\jygl 0.9267 &\jygl 0.9442 &\jygl 9.02 &\jygl &\jygl 0.7244 &\jygl {0.8024} &\jygl 54.07 &\jygl&\jygl 0.7018 &\jygl {0.7791} &\jygl 58.12\\
                        & S$^2$SAM              & 0.9228 & 0.9399 & 10.62 && 0.3588 & 0.4058 & 106.96 &&  0.3350 & 0.3689 & 106.76 \\
                        & \jygl S$^2$SAM + WCR     &\jygl 0.9294 &\jygl {0.9459} &\jygl {9.00} &\jygl &\jygl 0.5699 &\jygl 0.6818 &\jygl 72.46 &\jygl&\jygl 0.5447 &\jygl 0.6567 &\jygl 75.83\\
\bottomrule
\end{tabular}
\label{table:exp_seg}
\end{adjustbox}
\end{table*}

\subsection{Enhancing Pruning-Robust Optimizers in Vision Tasks}
\label{sec:vision}
\subsubsection{Image classification}
\label{sec:cls}
\noindent\textbf{CNN full training.}
Table~\ref{table:exp_cnn} reports CNN classification results on CIFAR-10, SVHN, and CIFAR-100 under one-shot pruning. Adding WCR to SGD, SFW, CrAM, SAM, and S$^2$SAM consistently improves post-pruning accuracy, with the largest gains appearing at high sparsity. At $96\%$ pruning on CIFAR-10, SAM improves from $48.36$ to $92.08$, while S$^2$SAM improves from $72.51$ to $93.44$. On SVHN, SAM improves from $9.74$ to $95.99$ at $96\%$ pruning. On CIFAR-100, S$^2$SAM improves from $1.13$ to $58.19$. We additionally provide an ablation study on SAM combined with DeepHoyer, $\ell_1$, and WCR under different regularization coefficients $\lambda$ in Appendix~\ref{append:ablation-sam}. These results suggest that WCR complements existing pruning-robust optimizers by improving the underlying weight distribution during training. 

\noindent\textbf{ViT fine-tuning.}
Table~\ref{table:exp_cnn} also reports ViT-B/32 results on CIFAR-100 and Tiny-ImageNet. WCR consistently improves transformer-based vision models under pruning, showing that the effect generalizes beyond CNNs. At $80\%$ pruning on CIFAR-100, SAM improves from $64.11$ to $84.95$, while on Tiny-ImageNet it improves from $41.57$ to $59.40$. Similar improvements are observed for CrAM and S$^2$SAM across pruning ratios.

\subsubsection{Segmentation results} \label{sec:seg} 
\begin{figure*}[t]
  \centering
  \includegraphics[width=1\linewidth]{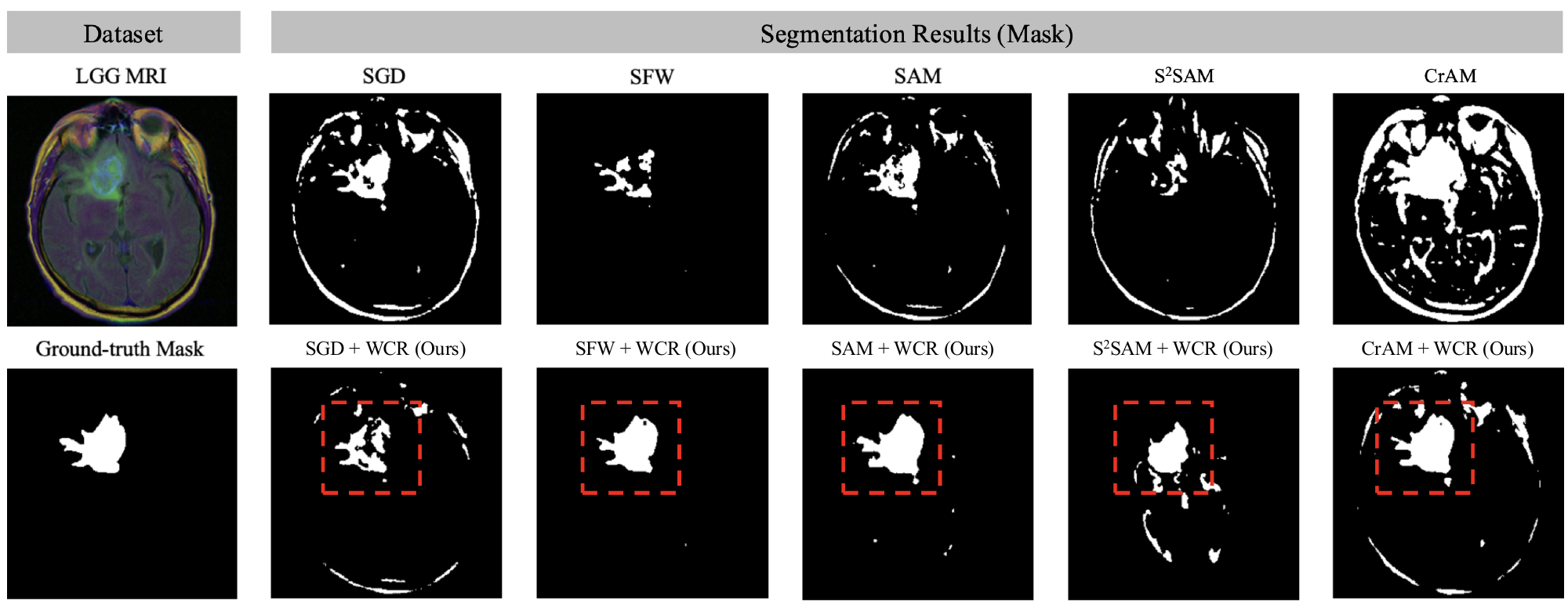}
  \caption{Qualitative segmentation on LGG MRI at 85\% pruning, with and without WCR. Additional examples are provided in Appendix~\ref{appendix:seg}.}
  \label{fig:tumor}
\end{figure*}

We further evaluate WCR in medical segmentation, where high sparsity can severely degrade dense predictions. Table~\ref{table:exp_seg} reports LGG Brain MRI segmentation with ResNet-50--UNet at Dense, $85\%$, and $88\%$ pruning. WCR consistently improves F1 and Tversky scores and reduces Hausdorff distance after pruning. At $85\%$ pruning, SAM improves from $0.3903$ to $0.7244$ F1 with WCR. At $88\%$ pruning, SGD recovers from failure to $0.5280$ F1. Figure~\ref{fig:tumor} qualitatively shows that WCR better preserves lesion regions, while baselines often produce fragmented or empty masks.

\begin{table*}[t]
\centering
\scriptsize
\caption{Accuracy (\%) on eight commonsense reasoning benchmarks for Qwen-2.5-1.5B and LLaMA-3.2-1B fine-tuned with LoRA on Commonsense-170K, under one-shot magnitude pruning at dense, 20\%, and 30\% pruning ratios.}
\vspace{3pt}
\begin{adjustbox}{width=1\columnwidth}
\begin{tabular}{l l l r r r r r r r r r}
\toprule
\textbf{Model} & \textbf{Pruning}&\textbf{Method} & \textbf{BoolQ} & \textbf{PIQA} & \textbf{SIQA} & \textbf{ARC-c} & \textbf{ARC-e} & \textbf{OBQA} & \textbf{HellaS} & \textbf{WinoG} & \textbf{AVG} \\
\midrule
\multirow{12}{*}{\begin{sideways}Qwen-2.5-1.5B\end{sideways}}
&\multirow{4}{*}{Dense} & AdamW  & 43.91 & 81.99 & 72.47 & 69.45 & 83.33 & 81.00 & 80.04 & 68.82 & 72.63\\
&& AdamW + $\ell_1$     & 65.05 & 80.90 & 74.05 & 65.44 & 79.88 & 75.60 & 75.80 & 73.40 & 73.77 \\
&& AdamW + DeepHoyer     & 39.08 & 83.73 & 75.54 & 74.49 & 89.02 & 83.60 & 87.30 & 74.66 & 75.93\\
&& \jygl AdamW + WCR     & \jygl 65.96 & \jygl 80.85 & \jygl 74.41 & \jygl 69.37 & \jygl 85.35 & \jygl 79.60 & \jygl 84.05 & \jygl 76.01 & \jygl 76.95\\
\cmidrule{2-12}
&\multirow{4}{*}{20\%} & AdamW          & 62.23 & 56.96 & 64.33 & 69.45 & 84.09 & 79.80 & 82.87 & 73.24 & 71.62\\
&& AdamW + $\ell_1$     & 64.43 & 78.78 & 73.23 & 63.48 & 78.16 & 76.40 & 73.03 & 71.19 & 72.34\\
&& AdamW + DeepHoyer     & 37.83 & 81.34 & 75.33 & 72.87 & 87.04 & 83.20 & 85.44 & 72.06 & 74.39\\
&& \jygl AdamW + WCR     & \jygl 65.78 & \jygl 80.25 & \jygl 73.90 & \jygl 68.09 & \jygl 83.75 & \jygl 77.00 & \jygl 81.22 & \jygl 72.45 & \jygl 75.31\\
\cmidrule{2-12}
&\multirow{4}{*}{30\%} & AdamW          & 62.17 & 46.68 & 62.23 & 63.91 & 80.39 & 74.20 & 78.18 & 64.33 & 66.51 \\
&& AdamW + $\ell_1$     & 63.09 & 75.19 & 67.81 & 57.68 & 71.42 & 70.00 & 67.31 & 67.32 & 67.48\\
&& AdamW + DeepHoyer     & 60.76 & 59.19 & 67.09 & 62.46 & 81.19 & 75.40 & 61.06 & 64.88 & 66.50\\
&& \jygl AdamW + WCR     & \jygl 65.66 & \jygl 71.06 & \jygl 71.44 & \jygl 60.24 & \jygl 76.89 & \jygl 70.40 & \jygl 70.85 & \jygl 69.77 & \jygl 69.54\\
\midrule
\multirow{12}{*}{\begin{sideways}LLaMA-3.2-1B\end{sideways}}
&\multirow{4}{*}{Dense} & AdamW          & 64.01 & 76.50 & 72.98 & 52.05 & 71.46 & 68.00 & 66.19 & 68.11 & 67.41\\
&& AdamW + $\ell_1$    & 60.95 & 66.00 & 67.35 & 49.23 & 65.45 & 62.20 & 52.44 & 64.88 & 61.06\\
&& AdamW + DeepHoyer    & 63.33 & 75.90 & 72.88 & 52.82 & 70.96 & 67.20 & 68.46 & 70.17 & 67.71\\
&& \jygl AdamW + WCR    & \jygl 62.35 & \jygl73.72 & \jygl70.57 & \jygl50.09 & \jygl67.34 & \jygl64.40 & \jygl60.75 & \jygl67.40 & \jygl64.58\\
\cmidrule{2-12}
&\multirow{4}{*}{20\%} & AdamW          & 61.44 & 67.74 & 66.17 & 42.06 & 58.16 & 54.60 & 48.70 & 61.56 & 57.55\\
&& AdamW + $\ell_1$    & 50.64 & 29.76 & 51.64 & 43.86 & 56.52 & 44.00 & 39.15 & 53.59 & 46.14\\
&& AdamW + DeepHoyer    & 61.35 & 71.76 & 70.37 & 48.38 & 64.14 & 62.60 & 53.25 & 64.01 & 61.98\\
&& \jygl AdamW + WCR    & \jygl62.20 & \jygl73.72 & \jygl69.19 & \jygl46.08 & \jygl63.01 & \jygl60.00 & \jygl59.97 & \jygl64.56 & \jygl62.34\\
\cmidrule{2-12}
&\multirow{4}{*}{30\%} & AdamW          & 60.55 & 55.60 & 52.30 & 32.59 & 41.12 & 23.20 & 26.32 & 52.96 & 43.08\\
&& AdamW + $\ell_1$    & 49.79 & 29.54 & 11.16 & 10.67 & 12.08 &  7.40 & 24.95 & 51.22 & 24.60\\
&& AdamW + DeepHoyer    & 57.52 & 55.39 & 49.64 & 30.97 & 47.22 & 45.60 & 11.01 & 56.12 & 44.18\\
&& \jygl AdamW + WCR    & \jygl61.47 & \jygl63.06 & \jygl59.98 & \jygl32.08 & \jygl47.73 & \jygl46.60 & \jygl10.40 & \jygl59.59 & \jygl47.61\\
\bottomrule
\end{tabular}
\end{adjustbox}
\label{tab:commonsense}
\end{table*}

\subsection{LLM Fine-Tuning: Comparison with Weight-Shaping Regularizers}
\label{sec:llm}
We evaluate WCR in LoRA-based LLM fine-tuning under a fixed AdamW + LoRA
pipeline on Qwen-2.5-1.5B and LLaMA-3.2-1B fine-tuned on Commonsense-170K,
and compare against $\ell_1$ and DeepHoyer regularization.
The regularization coefficients are selected through a $\lambda$ sweep on
Qwen-2.5-1.5B and then applied to LLaMA-3.2-1B. Full tuning results are
provided in Appendix~\ref{llm-hyper}.

Table~\ref{tab:commonsense} reports per-task and average accuracy under
one-shot magnitude pruning at Dense, $20\%$, and $30\%$ pruning ratios.
Across both models, WCR remains competitive after pruning and achieves the
highest average accuracy at $30\%$ pruning among the compared regularizers.
These results suggest that encouraging concentrated weight distributions
during LoRA fine-tuning can improve robustness under one-shot magnitude
pruning.

\subsection{Generalization analysis}
\label{generalization}
\begin{figure*}[t]
  \centering
  \includegraphics[width=1\linewidth]{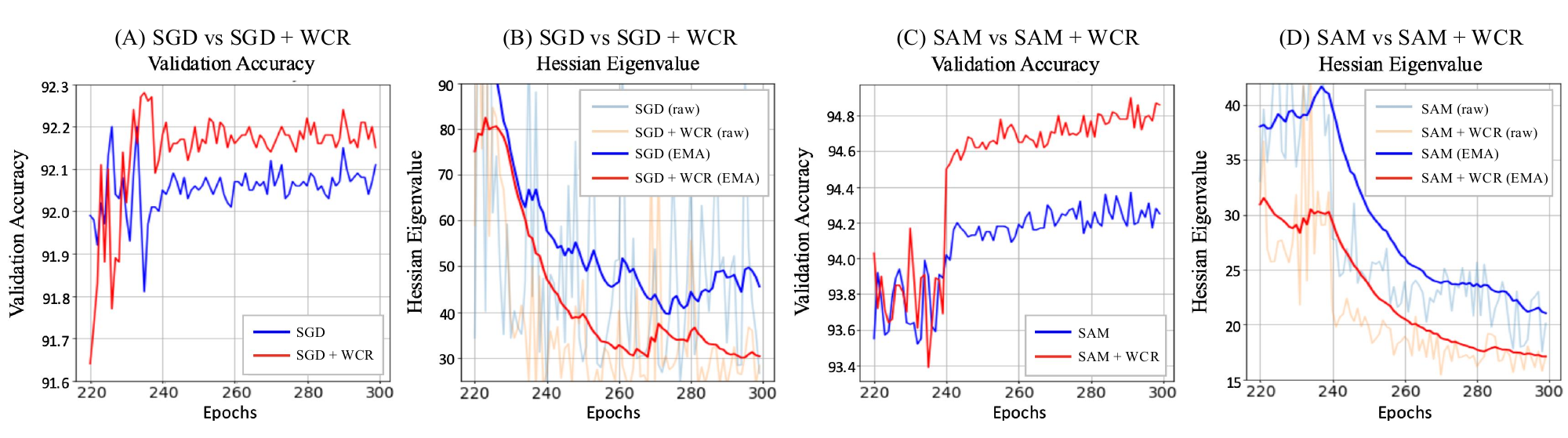}
  \caption{SGD/SAM with and without WCR on CIFAR-10/ResNet-18 (batch 1024, 300 epochs). (A,C) accuracy; (B,D) principal Hessian eigenvalue. WCR lowers curvature and raises accuracy.}
  \label{fig:generalization}
\end{figure*}

We measure the principal Hessian eigenvalue, a standard proxy for loss-surface curvature~\citep{lee2023achieving,keskar2016large}. 
Figure~\ref{fig:generalization} shows that WCR lowers this curvature for both SGD and SAM on CIFAR-10/ResNet-18 while improving validation accuracy, suggesting that WCR is associated with flatter solutions in this setting.
We use a longer training schedule (300 epochs) for this analysis to better assess generalization behavior and stability.

\section{Discussion}
Our study shows that encouraging weight concentration improves pruning robustness by changing the weight magnitude distribution during training. WCR concentrates weight magnitude onto a small subset of parameters while driving the rest toward near-zero values, creating a clearer separation between large- and small-magnitude weights. This allows magnitude pruning to remove many low-magnitude parameters with less performance degradation, leading to improved accuracy retention under high sparsity.

\noindent\textbf{Limitations and future work.}
Our theoretical analysis relies on a smoothness assumption on the concentration objective, and extending the analysis beyond this setting remains an important direction. Nevertheless, across a wide range of models and datasets, we empirically observe stable training behavior without noticeable instability. Our LLM experiments focus on the standard AdamW + LoRA pipeline and do not include SAM-family optimizers, which are not commonly used in parameter-efficient fine-tuning due to their additional overhead. However, our vision results demonstrate that WCR composes effectively with SAM, CrAM, and S$^2$SAM, especially at high sparsity. Extending this compositional analysis to LLM-scale models is an important direction. Additional future work includes structured pruning and evaluating robustness under distribution shifts.

\section{Conclusion}

This work introduces the Weight Concentration Regularizer (WCR), a training-time regularizer that improves pruning robustness without modifying standard optimization pipelines. WCR encourages a weight distribution in which a small subset of parameters carries most of the weight energy while the remaining weights are driven toward near-zero values, producing a clearer separation between large- and small-magnitude parameters. Under such a distribution, magnitude-based pruning can remove many low-magnitude parameters with reduced performance degradation. Empirically, WCR consistently improves pruning robustness across image classification, medical segmentation, and LoRA-based LLM fine-tuning, while composing effectively with existing pruning-robust optimizers such as SAM, CrAM, and S$^2$SAM. Overall, our results suggest that shaping the weight magnitude distribution during training is an effective approach for improving robustness under high sparsity.

\bibliography{example_paper}
\bibliographystyle{plain}

\clearpage
\newpage
\appendix

\clearpage
\onecolumn
\setcounter{page}{1}

\appendix
\renewcommand{\thesection}{\Alph{section}}

\section*{Appendix}
This appendix provides additional materials that complement the main paper. 
It includes the full theoretical analysis supporting the convergence behavior of our Weight Concentration Regularizer (WCR), extended implementation details for all experiments, and additional qualitative and quantitative results that further illustrate the effect of WCR.
The appendix is organized as follows:
\begin{itemize}
    \item \textbf{Appendix}~\ref{appendix-theory}: Theoretical analysis and proofs for the proposed WCR.
    \item \textbf{Appendix}~\ref{appendix-method}: PyTorch implementation of the Weight Concentration Regularizer.
    \item \textbf{Appendix}~\ref{appendix-expsetup}: Experimental setup, implementation details, and evaluation metrics.
    \item \textbf{Appendix}~\ref{appendix-exp}: Extended experimental results.
\end{itemize}

\section{Theoretical Analysis}
\label{appendix-theory}
\setcounter{theorem}{0}
In this section, we provide the detailed proofs of the lemmas and corollaries presented in the main paper. Specifically, we establish that adding the Weight Concentration Regularizer (WCR) preserves the convergence properties of stochastic gradient descent under standard smoothness and bounded-variance assumptions. \\

\begin{assumption}[$\beta_1$-smoothness]
\label{append:smoothness}
The loss function \(L:\mathbb{R}^d \to \mathbb{R}\) is \(\beta_1\)-smooth, i.e., its gradient is \(\beta_1\)-Lipschitz continuous $\|\nabla L(u) - \nabla L(v)\| \leq \beta_1 \|u-v\|, \forall u,v \in \mathbb{R}^d.$
Equivalently, for any \(u,v \in \mathbb{R}^d\), the following descent lemma holds:
\begin{align}
L(u) \leq L(v) + \langle \nabla L(v), u-v \rangle + \tfrac{\beta_1}{2}\|u-v\|^2.
\end{align}
\end{assumption}

\begin{assumption}[Unbiased stochastic gradient]
\label{append:unbiased}
At each iteration \(t\), the stochastic gradient \(\nabla L_t(w_t)\) is an unbiased estimator of the true gradient:
\begin{align}
\mathbb{E}[\nabla L_t(w_t)] = \nabla L(w_t).
\end{align}
\end{assumption}

\begin{assumption}[Local smoothness of $\psi(w)$]
\label{append:Bsmoothness}
The regularizer $\psi:\mathbb{R}^d \to \mathbb{R}$ is continuously differentiable. 
Moreover, we assume that $\psi(w)$ has a $\beta_2$-Lipschitz continuous gradient over a bounded domain $\mathcal{W}$ containing the iterates $\{w_t\}$, i.e.,
\begin{align}
\|\nabla \psi(w_1) - \nabla \psi(w_2)\| \le \beta_2 \|w_1 - w_2\|, \quad \forall w_1, w_2 \in \mathcal{W}.
\end{align}
\end{assumption}


We now provide the proofs for all the proposed lemmas and the theorem as follows.

\begin{lemma}
Under the $\beta_{1}$-smoothness of $L$ and the $\beta_{2}$-smoothness of $\psi$, 
the combined objective $L_{\mathrm{total}}(w)=L(w)+\lambda\psi(w)$
is $\beta$-smooth over $\mathcal{W}$ for $\beta \le \beta_{1}+\lambda\beta_{2}$.
Together with the bounded-variance conditions 
$\mathbb{E}[\nabla L_t(w_t)] = \nabla L(w_t)$ and 
$\mathbb{E}\!\left[\|\nabla L_t(w_t) - \nabla L(w_t)\|^2\right] \le \frac{\sigma^2}{b}$ (where the regularizer $\psi$ is deterministic, implying $\sigma_{\psi} = 0$),
if the step size satisfies $\eta \le 1/\beta$, then the classic SGD update
$w_{t+1} = w_t - \eta\big(\nabla L_t(w_t) + \lambda\nabla\psi(w_t)\big)$
guarantees:
\label{append:lemma1}
\begin{align}
\mathbb{E}\big[L&(w_{t+1})+\lambda \psi(w_{t+1})\big] \leq \mathbb{E}\big[L(w_t)+\lambda \psi(w_t)\big] - \frac{\eta}{2} \mathbb{E}[\|\nabla L(w_t)+\lambda\,\nabla \psi(w_t)\|^2]  + \frac{\eta^2 \beta}{2b}\sigma^2 
\end{align}
\end{lemma}

\begin{proof}
From the \(\beta\)-smoothness of \(L_{total}\), we have:
\begin{align*}
L(w_{t+1})+&\lambda \psi(w_{t+1}) \\ &\leq L(w_t)+\lambda \psi(w_t) + \mathbb{E}\!\left[\left\langle \nabla L(w_t)+\lambda \nabla \psi(w_t),\, w_{t+1} - w_t \right\rangle\right] + \frac{\beta}{2} \|w_{t+1} - w_t\|^2
\end{align*}

\noindent
Substituting the SGD update \(w_{t+1} = w_t - \eta (\nabla L(w) + \lambda \nabla \psi(w) )\), we obtain:
\begin{align*}
L(w_{t+1})&+\lambda \psi(w_{t+1})\\
&\leq L(w_t)+\lambda \psi(w_t)  - \eta \left\langle \nabla L(w_t)+\lambda \nabla \psi(w_t),\, \nabla L(w_t)+\lambda \nabla \psi(w_t) \right\rangle \\
& \quad + \frac{\eta^2 \beta}{2} \|L_{t}(w_t)+\lambda \psi(w_t)\|^2
\end{align*}

\noindent
Taking the expectation, and noting that $\mathbb{E}[\nabla L_t(w_t)] = \nabla L(w_t)$ and then $\mathbb{E}[\nabla L_t(w_t) + \lambda \nabla \psi(w)] = \nabla L(w_t)+\lambda\nabla \psi(w_t)$, we get:
\begin{align*}
\mathbb{E}\big[L(w_{t+1})+\lambda & \psi(w_{t+1})\big] \\
&\leq \mathbb{E}\big[L(w_t)+\lambda \psi(w_t)\big] - \eta \,\mathbb{E}[\left\langle \nabla L(w_t)+\lambda \nabla \psi(w_t),\, \nabla L(w_t)+\lambda \nabla \psi(w_t) \right\rangle]  \\
&\quad + \frac{\eta^2 \beta}{2} \mathbb{E}[\|\nabla L_{t}(w_t)+\lambda \nabla \psi(w_t)\|^2]  \\
&= \mathbb{E}\big[L(w_t)+\lambda \psi(w_t)\big] - \eta \,\mathbb{E}[\|\nabla L(w_t)+\lambda \nabla \psi(w_t)\|^2] \\
&\quad + \frac{\eta^2 \beta}{2} \mathbb{E}[\|\nabla L_{t}(w_t)+\lambda \nabla \psi(w_t)\|^2]
\end{align*}
The stochastic gradient variance is bounded as:
\begin{align*}
\mathbb{E}[L(w_{t+1})&+\lambda \psi(w_{t+1})] \\
&\leq \mathbb{E}[L(w_{t})+\lambda \psi(w_{t})]  
- \eta \mathbb{E}[\langle (\nabla L(w_{t})+\lambda \nabla\psi(w_{t})), (\nabla L_t(w_{t})+\lambda \nabla\psi(w_{t})) \rangle] \\
& \quad+ \frac{\eta^2 \beta}{2} \mathbb{E}[\|(\nabla L_t(w_{t})+\lambda \nabla\psi(w_{t}))\|^2]  \\
&= \mathbb{E}[(L(w_{t})+\lambda \psi(w_{t}))] - \eta \mathbb{E}[\|(\nabla L(w_{t})+\lambda \nabla\psi(w_{t}))\|^2] \\
&\quad + \frac{\eta^2 \beta}{2} \mathbb{E}[\|(\nabla L_t(w_{t})+\lambda \nabla\psi(w_{t}))\|^2]
\end{align*}
Expanding the variance term, we obtain
\begin{align}
\mathbb{E}&\big[\|\nabla L_{t}(w_t) + \lambda \nabla \psi(w_t)\|^2\big]  \\
&= \mathbb{E}\big[\| (\nabla L_t(w_t)+\lambda \nabla\psi(w_t)) 
      - (\nabla L(w_t)+\lambda \nabla\psi(w_t)) \nonumber 
      + (\nabla L(w_t)+\lambda \nabla\psi(w_t))\|^2\big]  \nonumber\\
&= \mathbb{E}\big[\|\nabla L_t(w_t)-\nabla L(w_t)\|^2\big] + 2\,\mathbb{E}\big[\langle \nabla L_t(w_t)-\nabla L(w_t),\,
      \nabla L(w_t)+\lambda\nabla\psi(w_t)\rangle\big] \nonumber\\
&\quad + \mathbb{E}\big[\|\nabla L(w_t)+\lambda\nabla\psi(w_t)\|^2\big]
\label{eq23}
\end{align}
By the unbiasedness assumption \eqref{eq23} simplifies to
\begin{align}
\mathbb{E}\big[\|\nabla L_{t}(w_t) + \lambda \nabla \psi(w_t)\|^2\big] 
= \mathbb{E}\big[\|\nabla L_t(w_t)-\nabla L(w_t)\|^2\big] 
  + \mathbb{E}\big[\|\nabla L(w_t)+\lambda\nabla\psi(w_t)\|^2\big]
\label{eq23-simplified}
\end{align}

\noindent
By the gradient-variance bound assumption, we have
\begin{align}
&\mathbb{E}\big[\|(\nabla L_{t}(w_t) + \lambda \nabla \psi(w_t)) 
      - (\nabla L(w_t) + \lambda \nabla \psi(w_t))\|^2\big] \\
&= \mathbb{E}\big[\|\nabla L_t(w_t)-\nabla L(w_t)\|^2\big]
\le \frac{\sigma^2}{b}
\label{eq24}
\end{align}
Combining \eqref{eq23-simplified} and \eqref{eq24}, we obtain
\begin{align}
\mathbb{E}\big[\|\nabla L_{t}(w_t) + \lambda \nabla \psi(w_t)\|^2\big] 
\le 
\mathbb{E}\big[\|\nabla L(w_t)+\lambda \nabla \psi(w_t)\|^2\big] 
+ \frac{\sigma^2}{b} 
\end{align}
Substituting the variance bound \eqref{eq23-simplified}–\eqref{eq24} into the descent inequality yields:
\begin{align*}
\mathbb{E}[L(w_{t+1}) & + \lambda \psi(w_{t+1})] \\
&\leq \mathbb{E}[L(w_{t}) + \lambda \psi(w_{t})] - \eta \mathbb{E}[\|\nabla L(w_{t}) + \lambda \nabla \psi(w_{t})\|^2] \\
& \quad + \frac{\eta^2 \beta}{2} \left( \mathbb{E}[\|\nabla L(w_{t}) + \lambda \nabla \psi(w_{t})\|^2] + \frac{\sigma^2}{b} \right)  \\
&= \mathbb{E}[L(w_{t}) + \lambda \psi(w_{t})] - \eta \mathbb{E}[\|\nabla L(w_{t}) + \lambda \nabla \psi(w_{t})\|^2] \\
& \quad + \frac{\eta^2 \beta}{2} \mathbb{E}[\|\nabla L(w_{t}) + \lambda \nabla \psi(w_{t})\|^2] + \frac{\eta^2 \beta\sigma^2}{2b}  \\
&= \mathbb{E}[L(w_{t}) + \lambda \psi(w_{t})] + \left( -\eta + \frac{\eta^2 \beta}{2} \right) \mathbb{E}[\|\nabla L(w_{t}) + \lambda \nabla \psi(w_{t})\|^2]  + \frac{\eta^2 \beta \sigma^2}{2b}  \\
&= \mathbb{E}[L(w_{t}) + \lambda \psi(w_{t})] - \left( \eta - \frac{\eta^2 \beta}{2} \right) \mathbb{E}[\|\nabla L(w_{t}) + \lambda \nabla \psi(w_{t})\|^2]  + \frac{\eta^2 \beta \sigma^2}{2b}
\end{align*}

\noindent
Finally, given \(\eta \leq \frac{1}{\beta}\), we have $\eta - \frac{\eta^2 \beta}{2} = \eta \left( 1 - \frac{\eta \beta}{2} \right) \geq \frac{\eta}{2}$ and, since \(\eta \beta \leq 1\) implies \(1 - \frac{\eta \beta}{2} \geq \frac{1}{2}\). 

Therefore,
\begin{align*}
\mathbb{E}\big[L(w_{t+1})+\lambda & \psi(w_{t+1})\big] \leq \mathbb{E}\big[L(w_t)+\lambda \psi(w_t)\big] - \frac{\eta}{2} \mathbb{E}[\|\nabla L(w_t)+\lambda\,\nabla \psi(w_t)\|^2] + \frac{\eta^2 \beta}{2b}\sigma^2
\end{align*}
\end{proof}

\begin{theorem}
Under the same smoothness and bounded-variance conditions stated in Lemma~\ref{append:lemma1}, and for a fixed step size $0 < \eta \le 1/\beta$, the SGD update satisfies:
\begin{align*}
\frac{1}{T} \sum_{t=0}^{T-1} &\mathbb{E}[\|\nabla L(w_t) + \lambda \nabla \psi(w_t)\|^2] \\
&\leq \frac{2}{T \eta} \big( (L(w_0) + \lambda \psi(w_0))\big) - \frac{2}{T \eta} \big(\mathbb{E}[(L(w_T) + \lambda \psi(w_T))] \big)  + \frac{\eta \beta \sigma^2}{2b}
\end{align*}
\end{theorem}

\begin{proof}
Based on the descent lemma for classic SGD, we have:
\begin{align}
\mathbb{E}[L(w_{t+1}) + \lambda  \psi(w_{t+1})] \leq \mathbb{E}[L(w_t) + \lambda \psi(w_t)] - \frac{\eta}{2} \mathbb{E}[\|\nabla L(w_t) + \lambda \nabla \psi(w_t)\|^2] + \frac{\eta^2 \beta \sigma^2}{2b} \label{eq29}
\end{align}
Averaging (\ref{eq29}) across \(T\) iterations, we obtain:
\begin{align*}
\frac{1}{T} &\sum_{t=0}^{T-1} \mathbb{E}[L(w_{t+1}) + \lambda  \psi(w_{t+1})] 
\\ &\leq \frac{1}{T} \sum_{t=0}^{T-1} \left( \mathbb{E}[L(w_{t}) + \lambda \psi(w_{t})]\right) - \frac{\eta}{2T} \sum_{t=0}^{T-1} \mathbb{E}[\| \nabla L(w_{t}) + \lambda \nabla \psi(w_{t})\|^2]  +\frac{1}{T} \sum_{t=0}^{T-1}\left( \frac{\eta^2 \beta \sigma^2}{2b} \right)
\end{align*}
Rearranging the terms and substituting $\frac{1}{T} \sum_{t=0}^{T-1} \big( \mathbb{E}[L(w_t) + \lambda \psi(w_t)] - \mathbb{E}[L(w_{t+1}) + \lambda \psi(w_{t+1})] \big) = \frac{1}{T} \big( L(w_0) + \lambda \psi(w_0) - \mathbb{E}[L(w_t) + \lambda \psi(w_t)] \big)$, we get:
\begin{align*}
\frac{\eta}{2T} &\sum_{t=0}^{T-1} \mathbb{E} [\|\nabla L(w_t) + \lambda  \nabla \psi(w_t)\|^2] \\
&\leq \frac{1}{T} \sum_{t=0}^{T-1} \big( \mathbb{E}[(L(w_t) + \lambda \psi(w_t))] - \mathbb{E}[L(w_{t+1}) + \lambda \psi(w_{t+1})] \big) + \frac{\eta^2 \beta \sigma^2}{2b}  \\
 &= \frac{1}{T} \big( (L(w_0) + \lambda \psi(w_0)) - \mathbb{E}[L(w_T) + \lambda \psi(w_T)] \big) + \frac{\eta^2 \beta \sigma^2}{2b} 
\end{align*}

\noindent
Dividing both sides by \(\frac{\eta}{2}\), we obtain:
\begin{align*}
\frac{1}{T} &\sum_{t=0}^{T-1} \mathbb{E}[\|(\nabla L(w_t) + \lambda \nabla \psi(w_t))\|^2] \\
&\leq \frac{2}{T \eta} \big( (L(w_0) + \lambda  \psi(w_0)) - \mathbb{E}[L(w_T) + \lambda  \psi(w_T)] \big)
+ \frac{\eta \beta \sigma^2}{2b}  \label{finalbound_appendix}
\end{align*}
\end{proof}

\begin{corollary}
Under the same smoothness and bounded-variance assumptions stated in 
Lemma~\ref{append:lemma1}, and assuming further that $L_{\mathrm{total}}$ is bounded below by $L^{\inf}_{\mathrm{total}} > -\infty$, let the step size be constant $\eta = \tfrac{c}{\sqrt{T}}$ with $0 < c \le \tfrac{1}{\beta}$. Then the SGD iterates satisfy:
\begin{align}
\frac{1}{T} & \sum_{t=0}^{T-1}
\mathbb{E}[\big\| \nabla L(w_t) + \lambda \nabla \psi(w_t)\big\|^2] \\
& \le \frac{2\big((L(w_0)+\lambda\psi(w_0))-L_{\text{total}}^{\inf}\big)}{c}\,T^{-1/2} \nonumber  + \frac{\beta c}{2b}\sigma^2\,T^{-1/2}
\end{align}
\end{corollary}

\begin{proof}
Starting from Theorem 5,
\begin{align}
\frac{1}{T}\!&\sum_{t=0}^{T-1}\!
\mathbb{E}\!\left[\|\nabla L(w_t)+\lambda\nabla\psi(w_t)\|^2\right] \\
&\le
\frac{2}{T\eta}\!\left((L(w_0)+\lambda\psi(w_0))-
\mathbb{E}[L(w_T)+\lambda\psi(w_T)]\right) +\frac{\beta\eta}{2b}\sigma^2
\end{align}
Because \(L(w_T)+\lambda\psi(w_T)\ge L_{\text{total}}^{\inf}\),
define \(\Delta_0=(L(w_0)+\lambda\psi(w_0))-L_{\text{total}}^{\inf}\).
Substituting \(\eta \le \tfrac{c}{\sqrt{T}}\) yields
\begin{align}
\frac{1}{T}\!\sum_{t=0}^{T-1}\!
\mathbb{E}\!\left[\|\nabla L(w_t)+\lambda\nabla\psi(w_t)\|^2\right]
&\le
\frac{2\Delta_0}{(c/\sqrt{T})T}
+\frac{\beta c}{2b}\sigma^2T^{-1/2}
\nonumber\\
&=
\frac{2\Delta_0}{c}\,T^{-1/2}
+\frac{\beta c}{2b}\sigma^2\,T^{-1/2}
\end{align}
The overall convergence rate is 
\(\mathcal{O}(\frac{1}{\sqrt{T}})\).\\
\end{proof}

\begin{corollary}[Diminishing step size]
\label{append:diminishing}
Under the same smoothness and bounded-variance assumptions stated in Lemma~\ref{append:lemma1},
and assuming further that the total objective $L(w)+\lambda\psi(w)$ is bounded below by 
$L^{\inf}_{\mathrm{total}} > -\infty$, consider SGD with a diminishing step-size sequence $(\eta_t)_{t\ge0}$ satisfying:
\begin{align*}
0<\eta_t \le \frac{1}{\beta},\quad
\sum_{t=0}^\infty \eta_t = \infty,\quad
\sum_{t=0}^\infty \eta_t^2 < \infty
\end{align*}
Then the (step-size)–weighted average of squared gradients vanishes when $\nabla L_\mathrm{{total}}(w_t)=\nabla L(w_t)+\lambda\nabla \psi(w_t)$:
\begin{align*}
\lim_{T\to\infty}\frac{1}{\sum_{t=0}^{T-1}\eta_t}
\sum_{t=0}^{T-1}\eta_t\,\mathbb{E}\left[\|\nabla L_{\mathrm{total}}(w_t)\|^2\right] = 0
\end{align*}
In particular,
\begin{align*}
\liminf_{T\to\infty}\ \min_{0\le t\le T-1}\ \mathbb{E}\left[\|\nabla L_{\mathrm{total}}(w_t)\|^2\right]=0
\end{align*}
Therefore, there exists a subsequence along which the expected squared gradient converges to $0$.
\end{corollary}

\begin{proof}
From the \(\beta\)-smoothness of \(L_{total}\) and the SGD update $w_{t+1}=w_t-\eta_t (\nabla L_t(w_t) + \lambda \nabla \psi(w_t)))$, the same argument as in the lemma yields, for any $t$ with $\eta_t\le 1/\beta$,
\begin{align*}
\mathbb{E}\!\big[L(w_{t+1})\big]
\;\le\;
\mathbb{E}\!\big[L(w_t)\big]
-\frac{\eta_t}{2}\,\mathbb{E}\!\left[\|\nabla L(w_t)+\lambda\nabla \psi(w_t)\|^2\right]
+\frac{\beta\,\eta_t^2}{2b}\,\sigma^2
\end{align*}
Summing from $t=0$ to $T-1$ and telescoping gives
\begin{align*}
\frac{1}{2}\sum_{t=0}^{T-1}\eta_t\, \mathbb{E}[\|\nabla L(w_t)+\lambda \nabla \psi(w_t)\|^2] &\le
\mathbb{E}\!\big[L(w_0)\big]-\mathbb{E}\!\big[L(w_T)\big]
+\frac{\beta}{2b}\,\sigma^2 \sum_{t=0}^{T-1}\eta_t^2 \\
&\le
\big(L(w_0)-L^{\inf}\big)
+\frac{\beta}{2b}\,\sigma^2\sum_{t=0}^{T-1}\eta_t^2
\end{align*}
Divide both sides by $S_T:=\sum_{t=0}^{T-1}\eta_t$ to obtain
\begin{align*}
\frac{1}{2S_T}\sum_{t=0}^{T-1}\eta_t\,\mathbb{E}[\|\nabla L(w_t)+ &\lambda\nabla \psi(w_t)\|^2 ] \le \frac{(L(w_0) + \lambda \psi(w_0))-(L + \lambda \psi)^{\inf}}{S_T} +\frac{\beta}{2b}\,\sigma^2\frac{\sum_{t=0}^{T-1}\eta_t^2}{S_T}
\end{align*}
By the step-size conditions, $S_T\to\infty$ and $\sum_{t=0}^{T-1}\eta_t^2/S_T\to 0$ as $T\to\infty$.
Hence the right-hand side tends to $0$, which proves
\begin{align*}
\lim_{T\to\infty}\frac{1}{\sum_{t=0}^{T-1}\eta_t}
\sum_{t=0}^{T-1}\eta_t\,\mathbb{E}\!\left[\|\nabla L(w_t)+\lambda\nabla \psi(w_t)\|^2\right] \;=\; 0
\end{align*}
Finally, since $\min_{0\le t\le T-1} a_t \le \frac{\sum_{t=0}^{T-1}\eta_t a_t}{\sum_{t=0}^{T-1}\eta_t}$ for any nonnegative sequence $(a_t)$, we get
\begin{align*}
\min_{0\le t\le T-1}\ \mathbb{E}\!\left[\|\nabla L(w_t)+\lambda\nabla \psi(w_t)\|^2\right]
&\le
\frac{\sum_{t=0}^{T-1}\eta_t\,\mathbb{E}\!\left[\|\nabla L(w_t)+\lambda\nabla \psi(w_t)\|^2\right]}{\sum_{t=0}^{T-1}\eta_t}
\\ \qquad &\xrightarrow[T\to\infty]{}\;0
\end{align*}
which implies that even with the additional regularizer $\psi(w)$, the expected squared gradient is guaranteed to converge to $0$ as $T \to \infty$.
\end{proof}

\section{PyTorch Implementation}
\label{appendix-method}
\begin{figure}[h]
  \centering
  \includegraphics[width=1\linewidth]{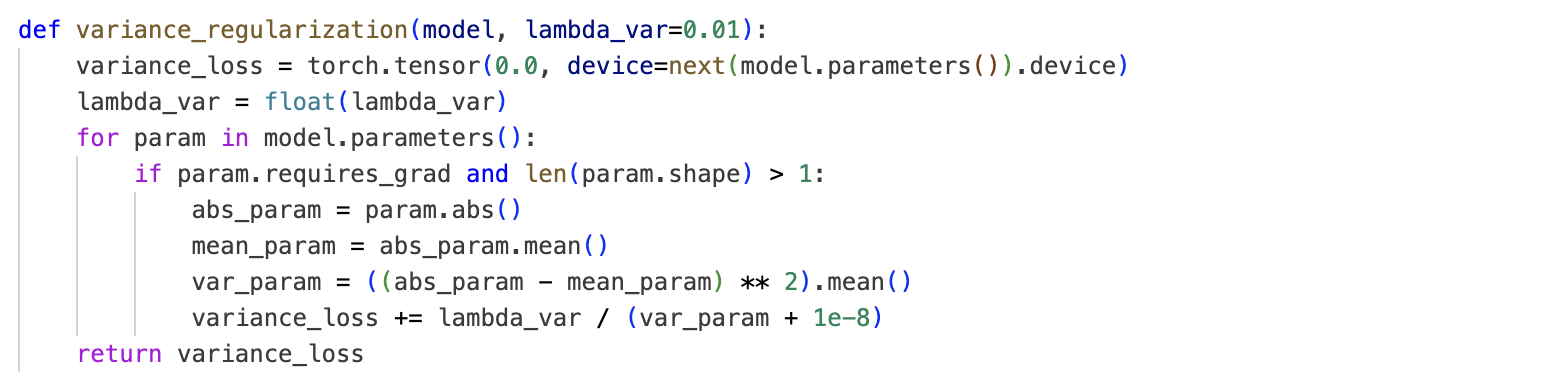}
  \caption{PyTorch implementation of the proposed Weight Concentration Regularizer (WCR).
For each layer, the regularizer computes the reciprocal of the variance of the
absolute weight values, concentrating weight energy onto a small subset of parameters to improve pruning robustness. The final regularization loss is the sum of these layer-wise terms.}
  \label{fig:sourcecode}
\end{figure}

Figure~\ref{fig:sourcecode} shows a PyTorch implementation of WCR designed for easy adoption. 
The regularizer is written as a standalone function that returns a single scalar loss term, so it can be directly added to any training objective as $L_{\text{total}}(w) = L_{\text{task}}(w) + \psi(w)$ without changing the model architecture or the optimizer. In practice, this makes WCR easy to plug into an existing PyTorch training loop with minimal code changes.

\section{Experimental Setups}
\label{appendix-expsetup}

\subsection{Implementation Details}
\label{appendix:exp}
All experiments were conducted on four NVIDIA A40 GPUs (48GB each). Each model was trained on a single GPU, while multiple runs were executed in parallel across the four devices. The full training configurations, including model architectures, datasets, number of epochs, batch sizes, and regularization coefficients, are summarized in Tables~\ref{tab:trainingsetting-append1}, \ref{tab:trainingsetting-append2}, and~\ref{tab:llm-training-append}.

\begin{table}[h]
\scriptsize
\caption{Summary of datasets, network architectures, and training settings used in vision experiments.}
\centering
\begin{tabular}{l|lllll} 
\toprule
\textbf{Task} & \textbf{Model} & \textbf{Dataset} & \textbf{\# Epochs} & \textbf{Batch Size} & \textbf{$\lambda$} \\
\midrule
\multirow{5}{*}{Classification} 
 & ResNet-18~\cite{resnet} & CIFAR-10~\cite{CIFAR} & 200 & 128 & 1e-5\\
 & ResNet-50~\cite{resnet} & SVHN~\cite{SVHN} & 200 & 128 & 1e-5\\
 & WideRes28-10~\cite{wideres2} & CIFAR-100~\cite{CIFAR} & 200 & 128 & 1e-5\\
 & ViT-B/32~\cite{VIT} & CIFAR-100~\cite{CIFAR} & 200 & 128 & 1e-2\\
 & ViT-B/32~\cite{VIT} & Tiny-ImageNet~\cite{tinyimagenet} & 200 & 128 & 1e-2\\
\midrule
Segmentation & Res50-Unet~\cite{unet} & LGG MRI~\cite{lgg} & 200 & 64 & 1e-5\\
\bottomrule
\end{tabular}
\label{tab:trainingsetting-append1}
\end{table}

\begin{table}[h]
\scriptsize
\caption{Training configurations for classification and segmentation experiments. ``--'' denotes settings that are not applicable.}
\centering
\begin{adjustbox}{width=1\columnwidth}
\begin{tabular}{l|lllllll} 
\toprule
\textbf{Task} & \textbf{Methods} & \textbf{Model} & \textbf{LR Initial} & \textbf{LR Scheduler} & \textbf{LR Decay Rate} & $\rho$ & \textbf{Momentum} \\
\midrule
\multirow{10}{*}{Classification} 
 & SGD & Res-18/50, WideRes-28 & 0.1 & Dynamic tuning & 0.7 & -- & 0.9 \\
 & SFW~\cite{sfw} & Res-18/50, WideRes-28 & 1.0 & Dynamic tuning & 0.7 & -- & 0.9 \\
 & CrAM~\cite{cram} & Res-18/50, WideRes-28 & 0.1 & Dynamic tuning & 0.7 & 0.05 & 0.9 \\
 & SAM~\cite{sam} & Res-18/50, WideRes-28 & 0.1 & Dynamic tuning & 0.7 & 0.5 & 0.9 \\
 & S$^2$SAM~\cite{ssam} & Res-18/50, WideRes-28 & 0.1 & Dynamic tuning & 0.7 & 0.5 & 0.9 \\
\cmidrule{2-8}
 & SGD & ViT-B/32 & 0.001 & Step decay & 0.5 & -- & 0.9 \\
 & SFW~\cite{sfw} & ViT-B/32 & -- & -- & -- & -- & -- \\
 & CrAM~\cite{cram} & ViT-B/32 & 0.001 & Step decay & 0.5 & 0.05 & 0.9 \\
 & SAM~\cite{sam} & ViT-B/32 & 0.001 & Step decay & 0.5 & 0.25 & 0.9 \\
 & S$^2$SAM~\cite{ssam} & ViT-B/32 & 0.001 & Step decay & 0.5 & 0.25 & 0.9 \\
\midrule
\multirow{5}{*}{Segmentation} 
 & SGD & Res50-Unet & 0.01 & Step decay & 0.5 & -- & 0.9 \\
 & SFW~\cite{sfw} & Res50-Unet & 0.01 & Step decay & 0.5 & -- & 0.9 \\
 & CrAM~\cite{cram} & Res50-Unet & 0.01 & Step decay & 0.5 & 0.05 & 0.9 \\
 & SAM~\cite{sam} & Res50-Unet & 0.01 & Step decay & 0.5 & 0.25 & 0.9 \\
 & S$^2$SAM~\cite{ssam} & Res50-Unet & 0.01 & Step decay & 0.5 & 0.25 & 0.9 \\
\bottomrule
\end{tabular}
\end{adjustbox}
\label{tab:trainingsetting-append2}
\end{table}

\begin{table}[h]
\footnotesize
\caption{LLM fine-tuning configuration used for Qwen-2.5-1.5B and LLaMA-3.2-1B on Commonsense-170K.}
\centering
\begin{tabular}{ll}
\toprule
Configuration & Value \\
\midrule
Fine-tuning method & LoRA \\
LoRA rank $r$ & 32 \\
LoRA $\alpha$ & 64 \\
LoRA dropout & 0.05 \\
Target modules & \texttt{q\_proj}, \texttt{k\_proj}, \texttt{v\_proj}, \texttt{o\_proj}, \texttt{gate\_proj}, \texttt{up\_proj}, \texttt{down\_proj} \\
Optimizer & AdamW \\
$\beta_1$, $\beta_2$ & 0.9, 0.999 \\
Weight decay & 0 \\
Learning rate & $2\times10^{-4}$ \\
LR schedule & Linear schedule with 100 warmup steps \\
Epochs & 2 \\
Batch size & 16 \\
Gradient accumulation & 4 \\
Cutoff length & 256 \\
Regularization coefficient $\lambda$ & $10^{-6}$ \\
Sweep values for $\lambda$ & $\{10^{-5}, 10^{-6}, 10^{-7}\}$ \\
Pruning ratios & Dense, 20\%, 30\% \\
\bottomrule
\end{tabular}
\label{tab:llm-training-append}
\end{table}

For classification tasks, we trained ResNet-18, ResNet-50, WideResNet-28-10, and ViT-B/32 for 200 epochs using a batch size of 128. Segmentation experiments on the LGG MRI dataset used ResNet-50--UNet for 200 epochs with a batch size of 64. For LLM experiments, we fine-tuned Qwen-2.5-1.5B and LLaMA-3.2-1B with LoRA on Commonsense-170K under the configuration in Table~\ref{tab:llm-training-append}. The regularization strength $\lambda$ was selected according to the model and task setting, as summarized in Tables~\ref{tab:trainingsetting-append1} and~\ref{tab:llm-training-append}.

\paragraph{Learning Rate Scheduling.}
The learning rate strategies differ across tasks:
\begin{itemize}
\item \textbf{Dynamic Tuning (CNN classification).}
The learning rate is reduced by a factor of 10 at one-third and two-thirds of the total training epochs. After epoch 20, the learning rate is additionally adjusted every five epochs using a moving-loss comparison: if the recent 5-epoch loss average exceeds the recent 10-epoch average, the learning rate is multiplied by $0.7$; otherwise it is slightly increased by multiplying by $1.06$.

\item \textbf{Step Decay (ViT and Segmentation).}
For ViT-B/32 and segmentation experiments, we use a step-decay rule, multiplying the learning rate by $0.5$ every 50 epochs.

\item \textbf{Linear Schedule (LLM fine-tuning).}
For LLM fine-tuning, we use a linear learning-rate schedule with 100 warmup steps.
\end{itemize}

All SFW experiments use the official default configuration provided in the public repository, including an initial learning rate of 1.0, dynamic\_change learning rate scheduling, momentum of 0.9, zero weight decay, a $k$-sparse polytope constraint with $K=10$ and $K_{\mathrm{frac}}=0.05$, diameter parameter $\tau=15$, constraint rescaling mode set to initialization, and the SFW\_Init option disabled. For constrained updates, SFW performs projection onto the $k$-sparse polytope at every optimization step. CrAM, SAM, and S$^{2}$SAM follow the default settings specified in their official repositories. However, for ViT-B/32 and ResNet50--UNet, the default perturbation radius $\rho=0.5$ in SAM and S$^{2}$SAM led to unstable training. We therefore selected a smaller $\rho$ by gradually reducing the value and choosing the one that yielded stable training and strong validation performance.

\subsection{Benchmark Datasets}
This work evaluates WCR across image classification, medical image segmentation, and LLM fine-tuning. All pruning is performed after training or fine-tuning without additional retraining.

\noindent \textbf{Image Classification.}
We assess pruning robustness on four classification benchmarks:
\begin{itemize}
    \item \textbf{CIFAR-10}~\cite{CIFAR}: A natural image dataset of $32\times32$ RGB images containing 10 object categories.
    \item \textbf{CIFAR-100}~\cite{CIFAR}: An extension of CIFAR-10 with 100 fine-grained categories.
    \item \textbf{SVHN}~\cite{SVHN}: A real-world digit classification dataset derived from street-view house numbers.
    \item \textbf{Tiny-ImageNet}~\cite{tinyimagenet}: A mid-scale ImageNet subset with $64\times64$ RGB images across 200 categories.
\end{itemize}

\noindent \textbf{Image Segmentation.}
We evaluate medical segmentation on:
\begin{itemize}
    \item \textbf{LGG Brain MRI Dataset}~\cite{lgg}: A collection of brain MRI scans from lower-grade glioma patients, paired with expert-annotated tumor masks for pixel-wise segmentation.
\end{itemize}

\noindent \textbf{LLM Fine-Tuning.}
We evaluate LoRA-based LLM fine-tuning on Commonsense-170K, following the protocol of DoRA~\cite{dora}. The training set is formed by combining the training splits of eight commonsense reasoning datasets: BoolQ, PIQA, SIQA, HellaSwag, WinoGrande, ARC-Easy, ARC-Challenge, and OBQA. Evaluation is performed on the corresponding test splits.

\subsection{Evaluation Metrics}
This section summarizes the evaluation metrics used across classification, segmentation, and LLM fine-tuning experiments.

\noindent \textbf{Image Classification.}
For all classification benchmarks, model performance is measured using validation accuracy.

\noindent \textbf{Image Segmentation.}
For segmentation on the LGG Brain MRI dataset, we use three complementary metrics:
\begin{itemize}
    \item \textbf{F1-score (F1)}: Measures the harmonic mean of precision and recall, capturing overlap between prediction and ground truth.
    \item \textbf{Tversky Index}: A generalization of the Dice score that asymmetrically penalizes false positives and false negatives.
    \item \textbf{Hausdorff Distance (H-Dist.)}: Quantifies the largest boundary deviation between the predicted mask and ground truth. Lower values indicate more accurate boundary localization.
\end{itemize}

\noindent \textbf{LLM Fine-Tuning.}
For LLM fine-tuning, we report test accuracy on each commonsense reasoning benchmark and the average accuracy across all eight tasks.

\subsection{One-shot Pruning Method}
All pruning is performed in a \textit{one-shot} manner after training or fine-tuning, without any retraining or additional fine-tuning.

For CNNs, convolutional weights are globally pruned using an $L_1$ magnitude threshold. For ViT-B/32, pruning is applied uniformly across the Query, Key, Value, Projection, and FFN layers. In additional appendix analyses, we also evaluate ViT models under pruning configurations that selectively prune only Q, QK, or QKV components in Section~\ref{append:ablation}.

For LLMs, we apply $L_1$ magnitude pruning to the LoRA-adapted attention and MLP projections after fine-tuning. Specifically, pruning is applied to \texttt{q\_proj}, \texttt{k\_proj}, \texttt{v\_proj}, \texttt{o\_proj}, \texttt{gate\_proj}, \texttt{up\_proj}, and \texttt{down\_proj}. We evaluate dense, 20\%, and 30\% pruned models.

\subsection{Principal Hessian Eigenvalue Estimation}
\label{app:hessian_eig}
We follow the Hessian eigenvalue computation procedure~\cite{hessian}. In particular, the top Hessian eigenvalues reported in Section~\ref{generalization} are computed using the power-iteration method with Hessian-vector products.

\begin{table}[h]
\centering
\scriptsize
\caption{Hyperparameters used for estimating the top Hessian eigenvalue $\widehat{\lambda}_{\max}$ via power iteration.}
\label{hhparams}
\begin{tabular}{llr}
\toprule
\textbf{Hyperparameter} & \textbf{Meaning} & \textbf{Value} \\
\midrule
Max iteration & Number of power-iteration steps $T$ & 20 \\
Hessian batch size & Mini-batch size $|\mathcal{B}|$ used for estimation & 1024 \\
\# of batch average & Number of mini-batches $M$ averaged to reduce noise & 12 \\
\# Epochs & Number of epochs & 300 \\
\bottomrule
\end{tabular}
\end{table}

We use the top Hessian eigenvalue $\widehat{\lambda}_{\max}$ as a curvature-based proxy to characterize the local sharpness of the loss landscape during training. By tracking $\widehat{\lambda}_{\max}$ across epochs, we evaluate whether WCR guides optimization toward flatter regions, supporting the generalization trends observed in Section~\ref{generalization}.

Let $w \in \mathbb{R}^d$ denote the vector of trainable parameters. For a mini-batch $\mathcal{B}$, let $\mathcal{L}_{\mathcal{B}}(w)$ denote the mini-batch objective, $g_{\mathcal{B}}(w):=\nabla_w \mathcal{L}_{\mathcal{B}}(w)$ the mini-batch gradient, and $H_{\mathcal{B}}(w):=\nabla_w^2 \mathcal{L}_{\mathcal{B}}(w)$ the Hessian.

We estimate $\widehat{\lambda}_{\max}$ by power iteration using Hessian-vector products. The mini-batch objective is
\begin{equation}
\mathcal{L}_{\mathcal{B}}(w)
=
\frac{1}{|\mathcal{B}|}\sum_{(x_i,y_i)\in\mathcal{B}} \ell(f_w(x_i),y_i).
\end{equation}
Starting from $v_0 \sim \mathcal{N}(0,I)$ normalized as $v_0 \leftarrow v_0/\|v_0\|_2$, we iterate for $t=0,1,\dots,T-1$:
\begin{align}
s_t(w) &= g_{\mathcal{B}}(w)^\top v_t, \label{eq:hess_scalar}\\
u_t(w) &= \nabla_w s_t(w)=H_{\mathcal{B}}(w)v_t, \label{eq:hvp}\\
\lambda_t &= \frac{v_t^\top u_t(w)}{v_t^\top v_t}
= \frac{v_t^\top H_{\mathcal{B}}(w)v_t}{\|v_t\|_2^2}, \label{eq:rayleigh}\\
v_{t+1} &= \frac{u_t(w)}{\|u_t(w)\|_2}. \label{eq:power_update}
\end{align}
We report $\widehat{\lambda}_{\max}:=\lambda_{T-1}$. To reduce mini-batch noise, we optionally average over $M$ batches:
\begin{equation}
\widehat{\lambda}_{\max}
=
\frac{1}{M}\sum_{m=1}^M \widehat{\lambda}_{\max}(\mathcal{B}_m).
\end{equation}

For the Hessian analysis in Section~\ref{generalization} and Figure~\ref{fig:generalization}, we use the setting in Table~\ref{hhparams}. All remaining experiments use the training configurations described in Section~\ref{appendix:exp}.


\newpage
\clearpage

\section{Experimental results}
\label{appendix-exp}
This section presents additional experimental results that complement the findings reported in the main paper. We first provide extended ablation studies examining the behavior of the proposed Weight Concentration Regularizer (WCR) under different regularization strengths, architectures, and pruning configurations.

\subsection{Additional results with aggressive pruning}
Table~\ref{table:exp_all_cnn-append} reports additional classification results conducted at pruning rates higher than those presented in the main paper. These experiments evaluate pruning robustness at extreme sparsity levels (96--98\%) on CIFAR-10 and SVHN using ResNet-18 and ResNet-50.

\begin{table*}[h]
\centering
\caption{Accuracy comparison (\%) across high pruning rates and datasets (CIFAR-10, SVHN) using ResNet-18/50, with and without WCR, denoted as ``w/o'' and ``w.''}
\vspace{3pt}
\footnotesize
\begin{tabular}{llc>{\columncolor{gray!15}}cc>{\columncolor{gray!15}}cc>{\columncolor{gray!15}}cc>{\columncolor{gray!15}}c}
\toprule
& &
\multicolumn{2}{c}{\textbf{Dense}} &
\multicolumn{2}{c}{\textbf{96\% Pruned}} &
\multicolumn{2}{c}{\textbf{97\% Pruned}} &
\multicolumn{2}{c}{\textbf{98\% Pruned}} \\
\cmidrule(lr){3-4} \cmidrule(lr){5-6} \cmidrule(lr){7-8} \cmidrule(lr){9-10}
\textbf{Setting} & \textbf{Method} &
\textbf{w/o} & \textbf{w} &
\textbf{w/o} & \textbf{w} &
\textbf{w/o} & \textbf{w} &
\textbf{w/o} & \textbf{w} \\
\midrule
\multirow{5}{*}{\shortstack[l]{ResNet-18\\CIFAR-10}}
 & SGD                       & 92.75 & 92.63 & 19.69 & 25.70 & 19.22 & 20.30 & 15.32 & 16.31 \\
 & SFW            & 92.54 & 92.34 & 84.15 & 89.93 & 59.12 & 66.52 & 12.61 & 15.02 \\
 & CrAM           & 94.19 & 93.86 & 66.66 & 81.72 & 35.00 & 69.08 & 17.27 & 41.77 \\
 & SAM             & 95.19 & 94.08 & 48.36 & 92.08 & 36.87 & 87.33 & 21.54 & 54.56 \\
 & S$^2$SAM       & 95.23 & 93.71 & 72.51 & 93.44 & 37.06 & 89.72 & 17.37 & 81.40 \\
\midrule
\multirow{5}{*}{\shortstack[l]{ResNet-50\\SVHN}}
 & SGD                       & 94.44 & 94.51 & 9.15  & 10.05 & 9.15  & 10.05 & 9.15  & 10.05 \\
 & SFW             & 95.79 & 95.80 & 18.20 & 95.49 & 6.69  & 9.69  & 6.69  & 9.69  \\
 & CrAM           & 95.97 & 95.79 & 27.78 & 31.26 & 11.34 & 12.86 & 6.87  & 8.47  \\
 & SAM             & 96.96 & 95.98 & 9.74  & 95.99 & 9.69  & 95.90 & 9.75  & 95.59 \\
 & S$^2$SAM       & 96.88 & 95.78 & 17.28 & 95.78 & 10.93 & 95.74 & 9.56  & 95.62 \\
\bottomrule
\end{tabular}
\label{table:exp_all_cnn-append}
\end{table*}

Across high-sparsity pruning settings, the proposed WCR consistently improves accuracy under very high pruning, even when baseline performance collapses. The gains are most prominent when coupled with sharpness-aware optimizers such as SAM and S$^2$SAM, where WCR preserves most of the dense-model performance despite removing up to 98\% of the weights.

\newpage
\clearpage

\subsection{Impact of Regularization Coefficient $\lambda$ for Vision tasks}
\label{append:ablation}

Tables~\ref{tab:lambda_tuning} and~\ref{tab:vit-lambda_tuning-append2} study
the effect of the regularization coefficient $\lambda$ on post-pruning accuracy
and the resulting weight statistics. Consistent with our analysis, increasing
$\lambda$ enlarges the overall weight variance $\mathrm{Var}(w)$, reflecting
that WCR concentrates weight energy onto a smaller subset of parameters and
thereby increases the spread between large and small weights.

For CNNs (ResNet-18 on CIFAR-10, Table~\ref{tab:lambda_tuning}), the optimal
$\lambda$ depends on the base optimizer. Under SGD, $\lambda=1\times10^{-5}$
gives the best accuracy at moderate pruning rates (90--94\%), while smaller
$\lambda=5\times10^{-6}$ becomes preferable at the most aggressive 96\% rate.
Under SAM, the implicit flatness bias already suppresses $\mathrm{Var}(w)$ by
roughly an order of magnitude (cf.\ the NA columns), and WCR remains effective
across a wide range of $\lambda \in [10^{-6}, 10^{-5}]$, recovering accuracy
from 48.36\% to over 92\% at 96\% pruning.

For ViTs (ViT-B/32 on CIFAR-100, Table~\ref{tab:vit-lambda_tuning-append2}),
a substantially larger $\lambda=1\times10^{-2}$ is required to obtain
meaningful pruning resilience across Q-, QK-, and QKV-pruning configurations.
The gap between optimal $\lambda$ values for CNNs and ViTs reflects the
different scales of attention weights relative to convolutional kernels;
overly large $\lambda$ (e.g., $1\times10^{-1}$) eventually hurts accuracy,
particularly under aggressive QK- and QKV-pruning where it interferes with
the attention structure itself.

\begin{table*}[h]
\centering
\caption{Performance comparison across different regularization coefficient ($\lambda$) values on CIFAR-10 with ResNet-18. The column labeled \textit{NA} denotes the baseline without the variance regularizer.}
\vspace{3pt}
\footnotesize
\begin{tabular}{l|llllllll}
\toprule
\multicolumn{9}{c}{\textbf{SGD}}\\
\midrule
Pruning $\quad \lambda$ & NA & 1e-6 & 5e-6 & 1e-5 & 5e-5 & 1e-4 & 5e-4 & 1e-3 \\
\cmidrule{2-9}
Rate $\qquad Var(w)$ & 0.0019 & 0.0021 & 0.0030 & 0.0035 & 0.0060 & 0.0081 & 0.0165 & 0.0226 \\
\midrule
Dense & 92.75 & 93.14 & 91.87 & 92.63 & 92.79 & 91.62 & 90.54 & 90.66 \\
90\%  & 43.57 & 68.55 & 77.00 & 90.88 & 76.73 & 64.87 & 66.22 & 58.22 \\
92\%  & 31.60 & 57.53 & 65.47 & 80.24 & 68.05 & 53.47 & 52.11 & 47.39 \\
94\%  & 26.10 & 41.02 & 51.0  & 59.91 & 49.56 & 41.62 & 31.5  & 35.25 \\
96\%  & 19.69 & 24.53 & 36.79 & 25.70 & 20.60 & 21.72 & 12.75 & 20.18 \\
\midrule
\multicolumn{9}{c}{\textbf{SAM}}\\
\midrule
Pruning $\quad \lambda$ & NA & 1e-6 & 5e-6 & 1e-5 & 5e-5 & 1e-4 & 5e-4 & 1e-3 \\
\cmidrule{2-9}
Rate $\qquad Var(w)$ & 0.00005 & 0.00010 & 0.00018 & 0.00024 & 0.00050 & 0.00069 & 0.00151 & 0.00212 \\
\midrule
Dense & 95.19 & 94.65 & 94.11 & 94.08 & 92.81 & 91.93 & 90.75 & 90.29 \\
90\%  & 91.26 & 94.47 & 94.22 & 94.04 & 92.77 & 91.82 & 90.75 & 90.13 \\
92\%  & 87.20 & 94.15 & 94.04 & 93.91 & 92.77 & 91.82 & 90.71 & 90.10 \\
94\%  & 76.29 & 92.61 & 93.69 & 93.77 & 92.69 & 91.74 & 90.69 & 90.08 \\
96\%  & 48.36 & 84.54 & 91.91 & 92.08 & 92.50 & 91.61 & 89.75 & 87.72 \\
\bottomrule
\end{tabular}

\label{tab:lambda_tuning}
\end{table*}

\begin{table*}[t]
\centering
\footnotesize
\caption{Performance comparison of ImageNet-Pretrained Vision Transformer (ViT-B/32) models on CIFAR-100 under different pruning configurations and regularization strengths $\lambda$. 
Each block shows results for Q, QK, and QKV-pruning, where pruning is applied respectively to the Query, Query–Key, and Query–Key–Value within the self-attention mechanism.
}
\vspace{3pt}
\begin{tabular}{llllllll}
\toprule
\textbf{Pruning }\quad $\lambda$ & \textbf{N/A} &\textbf{ 1e-5} & \textbf{1e-4} & \textbf{1e-3} & \textbf{1e-2} &\textbf{ 1e-1} \\
Rate \qquad $Var(w)$ & 0.0070 & 0.0070 & 0.0073 & 0.0101 & 0.0209 & 0.0574 \\
\midrule
\multicolumn{7}{c}{{Q-Pruning}} \\
\midrule
Dense & 89.84 & 90.00 & 90.15 & 89.62 & 88.75 & 85.23 \\
60\% & 89.52 & 89.75 & 89.70 & 89.43 & 88.65 & 85.23 \\
70\% & 88.88 & 89.15 & 89.19 & 89.04 & 88.58 & 85.16 \\
80\% & 87.57 & 87.63 & 87.95 & 88.15 & 87.90 & 84.69 \\
90\% & 80.29 & 81.27 & 81.51 & 83.84 & 85.13 & 82.75 \\
92\% & 76.25 & 77.44 & 77.68 & 81.22 & 83.04 & 81.22 \\
94\% & 68.97 & 70.52 & 70.66 & 76.48 & 78.88 & 78.54 \\
96\% & 56.31 & 57.86 & 58.25 & 66.83 & 70.33 & 71.73 \\
\midrule
\multicolumn{7}{c}{{QK-Pruning}} \\
\midrule
Dense & 89.84 & 90.00 & 90.15 & 89.62 & 88.75 & 85.23 \\
60\% & 88.75 & 89.25 & 89.11 & 89.29 & 88.79 & 85.23 \\
70\% & 87.09 & 87.23 & 87.47 & 88.29 & 88.11 & 85.15 \\
80\% & 79.52 & 80.72 & 81.65 & 84.90 & 86.06 & 84.16 \\
90\% & 51.54 & 52.61 & 54.57 & 65.31 & 73.41 & 75.01 \\
92\% & 41.56 & 43.22 & 43.84 & 54.82 & 64.43 & 50.61 \\
94\% & 31.67 & 33.52 & 33.45 & 41.31 & 50.65 & 33.18 \\
96\% & 23.32 & 24.75 & 24.06 & 27.64 & 33.25 & 19.52 \\
\midrule
\multicolumn{7}{c}{{QKV-Pruning}} \\
\midrule
Dense & 89.84 & 90.00 & 90.15 & 89.62 & 88.75 & 85.23 \\
60\% & 86.30 & 86.68 & 87.29 & 88.23 & 88.58 & 85.23 \\
70\% & 78.33 & 78.76 & 80.45 & 84.83 & 85.66 & 83.81 \\
80\% & 51.13 & 52.93 & 55.48 & 69.63 & 73.69 & 68.11 \\
90\% & 9.55 & 12.18 & 11.14 & 15.00 & 14.63 & 5.07 \\
92\% & 7.23 & 7.83 & 7.75 & 9.21 & 8.82 & 2.84 \\
94\% & 5.07 & 6.02 & 5.27 & 6.29 & 3.32 & 1.78 \\
96\% & 3.04 & 4.42 & 3.62 & 4.61 & 2.31 & 1.51 \\
\bottomrule
\end{tabular}
\label{tab:vit-lambda_tuning-append2}
\end{table*}

\clearpage
\newpage

\subsection{Ablation study for SAM with regularizers}
\label{append:ablation-sam}
To further analyze the interaction between pruning-robust optimization and weight-shaping regularization, we evaluate SAM combined with DeepHoyer, $\ell_1$, and WCR under different regularization coefficients $\lambda$. Table~\ref{tab:ablation_l1wcr} reports the validation accuracy of ResNet-18 on CIFAR-10 under one-shot magnitude pruning. The results show that the effectiveness of each regularizer depends on the choice of $\lambda$ and pruning ratio. In particular, WCR consistently maintains strong post-pruning accuracy across pruning levels and remains effective even at high sparsity.

\begin{table}[h]
\centering
\footnotesize
\caption{Validation accuracy (\%) of ResNet-18 on CIFAR-10 under one-shot magnitude pruning for SAM combined with DeepHoyer, $\ell_1$, and WCR, evaluated with different regularization coefficients $\lambda$.}
\begin{tabular}{
    lcccccccccccc
} 
\toprule
Pruning Rate & SAM & \multicolumn{3}{c}{\textbf{SAM + DeepHoyer}} &&\multicolumn{3}{c}{\textbf{SAM + $\ell_1$}} && \multicolumn{3}{c}{\textbf{SAM + WCR (Ours)}}\\
\cmidrule{3-5} \cmidrule{7-9} \cmidrule{11-13}
$\lambda$ & NA & 1e-4 & 1e-5 & 1e-6  && 1e-4 & 1e-5 & 1e-6 && 1e-4 & 1e-5 & 1e-6 \\
\midrule
Dense & 92.75 & 65.13 & 90.31 & 92.04 && 89.64 & 90.18 & 94.12  && 91.93 & 94.08 & 94.65 \\
90\% & 43.57 & 63.90 & 89.14 & 92.02 && 89.63 & 90.16 & 94.09  && 91.82 & 94.04 & 94.47\\
92\% & 31.60 & 63.54 & 89.03 & 91.43 && 88.16 & 90.08 & 93.75 && 91.82 & 93.91 & 94.15\\
94\% & 26.10 & 62.15 & 88.91 & 90.12 && 88.03 & 89.64 & 91.68 && 91.74 & 93.77 & 92.61\\
96\% & 19.69 & 62.07 & 88.89 & 90.05 && 73.59 & 80.56 & 82.79 && 91.61 & 92.08 & 84.54 \\
\bottomrule
\end{tabular}
\label{tab:ablation_l1wcr}
\end{table}

\clearpage
\newpage

\subsection{Impact of Regularization Coefficient $\lambda$ for LLM Fine-Tuning}
\label{llm-hyper}
Table~\ref{tab:qwen2.5-1.5b} reports the effect of the regularization
coefficient $\lambda$ on Qwen2.5-1.5B under LoRA fine-tuning followed by
one-shot magnitude pruning. We compare $\ell_1$, DeepHoyer, and WCR across
$\lambda \in \{10^{-7}, 10^{-6}, 10^{-5}\}$ under dense, $20\%$, and $30\%$
pruning.

WCR with $\lambda = 10^{-6}$ achieves the highest average accuracy at both
$20\%$ and $30\%$ pruning, reaching $75.31\%$ and $69.54\%$, respectively.
At dense, WCR with $\lambda = 10^{-7}$ attains the highest average accuracy
in the table ($77.60\%$). Across the three $\lambda$ values, WCR's
$30\%$-pruned average ranges from $64.76\%$ to $69.54\%$, $\ell_1$ from
$24.49\%$ to $67.48\%$, and DeepHoyer from $57.61\%$ to $66.50\%$. The
unregularized baseline reaches $72.63\%$, $71.62\%$, and $66.51\%$ at
dense, $20\%$, and $30\%$ pruning, respectively.

For $\ell_1$, the $30\%$-pruned average decreases from $67.48\%$ at
$\lambda = 10^{-7}$ to $49.81\%$ at $\lambda = 10^{-6}$ and $24.49\%$ at
$\lambda = 10^{-5}$. For DeepHoyer, the dense average is highest at
$\lambda = 10^{-6}$ ($75.93\%$); at $\lambda = 10^{-5}$, the $20\%$-pruned
average is $64.45\%$ and the $30\%$-pruned average is $65.45\%$.

\begin{table}[h]
\centering
\caption{Per-task test accuracy (\%) of Qwen2.5-1.5B with LoRA ($r{=}32$) under one-shot magnitude pruning at dense, 20\%, and 30\% pruning ratios, across different regularization methods and coefficients $\lambda$.}
\label{tab:qwen2.5-1.5b}
\resizebox{\linewidth}{!}{%
\begin{tabular}{llcccccccccc}
\toprule
\textbf{Reg} & $\lambda$ & \textbf{Pruning Rate} & \textbf{BoolQ} & \textbf{PIQA} & \textbf{SIQA} & \textbf{ARC-C} & \textbf{ARC-E} & \textbf{OBQA} & \textbf{HellaS} & \textbf{WinoG} & \textbf{Avg} \\
\midrule
\multirow{3}{*}{none}  & \multirow{3}{*}{N/A} & dense & 43.91 & 81.99 & 72.47 & 69.45 & 83.33 & 81.00 & 80.04 & 68.82 & 72.63\\
                       &                                       & 20\%  & 62.23 & 56.96 & 64.33 & 69.45 & 84.09 & 79.80 & 82.87 & 73.24 & 71.62\\
                       &                                       & 30\%  & 62.17 & 46.68 & 62.23 & 63.91 & 80.39 & 74.20 & 78.18 & 64.33 & 66.51 \\
\midrule
\multirow{9}{*}{l1}    & \multirow{3}{*}{$1\!\times\!10^{-7}$} & dense & 65.05 & 80.90 & 74.05 & 65.44 & 79.88 & 75.60 & 75.80 & 73.40 & 73.77 \\
                       &                                       & 20\%  & 64.43 & 78.78 & 73.23 & 63.48 & 78.16 & 76.40 & 73.03 & 71.19 & 72.34 \\
                       &                                       & 30\%  & 63.09 & 75.19 & 67.81 & 57.68 & 71.42 & 70.00 & 67.31 & 67.32 & 67.48 \\
                       \cmidrule{2-12}
                       & \multirow{3}{*}{$1\!\times\!10^{-6}$} & dense & 61.99 & 67.36 & 63.92 & 42.24 & 58.04 & 55.80 & 40.14 & 62.67 & 56.52 \\
                       &                                       & 20\%  & 62.08 & 65.45 & 62.28 & 39.68 & 56.40 & 56.40 & 37.39 & 62.12 & 55.22 \\
                       &                                       & 30\%  & 61.22 & 60.88 & 55.89 & 35.41 & 47.35 & 48.20 & 32.22 & 57.30 & 49.81 \\
                       \cmidrule{2-12}
                       & \multirow{3}{*}{$1\!\times\!10^{-5}$} & dense & 62.26 & 59.47 & 57.16 & 33.53 & 42.80 & 45.20 & 27.88 & 57.62 & 48.24 \\
                       &                                       & 20\%  & 62.17 & 53.16 & 56.96 & 32.08 & 40.66 & 45.00 & 25.83 & 57.38 & 46.65 \\
                       &                                       & 30\%  & 57.43 & 35.80 & 13.00 &  9.04 & 10.40 & 14.20 &  9.38 & 46.65 & 24.49 \\
\midrule
\multirow{9}{*}{DeepHoyer} & \multirow{3}{*}{$1\!\times\!10^{-7}$} & dense & 46.42 & 81.50 & 73.90 & 70.22 & 86.87 & 82.00 & 57.70 & 72.77 & 71.42 \\
                       &                                       & 20\%  & 62.20 & 71.93 & 68.27 & 63.31 & 81.73 & 74.00 & 52.06 & 57.62 & 66.39 \\
                       &                                       & 30\%  & 63.91 & 75.52 & 30.45 & 69.20 & 82.74 & 71.60 &  0.85 & 66.61 & 57.61 \\
                       \cmidrule{2-12}
                       & \multirow{3}{*}{$1\!\times\!10^{-6}$} & dense & 39.08 & 83.73 & 75.54 & 74.49 & 89.02 & 83.60 & 87.30 & 74.66 & 75.93 \\
                       &                                       & 20\%  & 37.83 & 81.34 & 75.33 & 72.87 & 87.04 & 83.20 & 85.44 & 72.06 & 74.39 \\
                       &                                       & 30\%  & 60.76 & 59.19 & 67.09 & 62.46 & 81.19 & 75.40 & 61.06 & 64.88 & 66.50 \\
                       \cmidrule{2-12}
                       & \multirow{3}{*}{$1\!\times\!10^{-5}$} & dense & 66.82 & 82.97 & 75.08 & 74.40 & 87.96 & 82.80 & 67.18 & 61.33 & 74.82 \\
                       &                                       & 20\%  & 66.76 & 83.19 & 74.56 & 71.25 & 87.12 & 80.40 & 28.18 & 24.15 & 64.45 \\
                       &                                       & 30\%  & 65.90 & 76.50 & 69.75 & 64.16 & 81.65 & 76.60 & 44.20 & 44.83 & 65.45 \\
\midrule
\multirow{9}{*}{WCR}   & \multirow{3}{*}{$1\!\times\!10^{-7}$} & dense & 64.53 & 81.66 & 74.62 & 72.27 & 86.91 & 81.60 & 85.01 & 74.19 & 77.60 \\
                       &                                       & 20\%  & 63.43 & 51.74 & 70.47 & 67.66 & 85.77 & 77.40 & 73.74 & 75.93 & 70.77 \\
                       &                                       & 30\%  & 63.82 & 77.26 & 45.60 & 55.63 & 74.16 & 70.00 & 75.46 & 56.12 & 64.76 \\
                        \cmidrule{2-12}
                        & \multirow{3}{*}{$1\!\times\!10^{-6}$} & dense & 65.96 & 80.85 & 74.41 & 69.37 & 85.35 & 79.60 & 84.05 & 76.01 & 76.95 \\
                       &                                       & 20\%  & 65.78 & 80.25 & 73.90 & 68.09 & 83.75 & 77.00 & 81.22 & 72.45 & 75.31 \\
                       &                                       & 30\%  & 65.66 & 71.06 & 71.44 & 60.24 & 76.89 & 70.40 & 70.85 & 69.77 & 69.54 \\
                       \cmidrule{2-12}
                       & \multirow{3}{*}{$1\!\times\!10^{-5}$} & dense & 65.12 & 79.84 & 73.62 & 67.81 & 83.92 & 78.20 & 82.01 & 74.38 & 75.61 \\
                       &                                       & 20\%  & 64.89 & 78.73 & 72.68 & 66.35 & 81.94 & 75.60 & 78.94 & 71.02 & 73.77 \\
                       &                                       & 30\%  & 64.71 & 69.32 & 69.87 & 58.47 & 74.83 & 68.80 & 68.24 & 68.15 & 67.80 \\
\bottomrule
\end{tabular}}
\end{table}

\clearpage
\newpage

\begin{figure*}[t]
  \centering
  \includegraphics[width=1\textwidth]{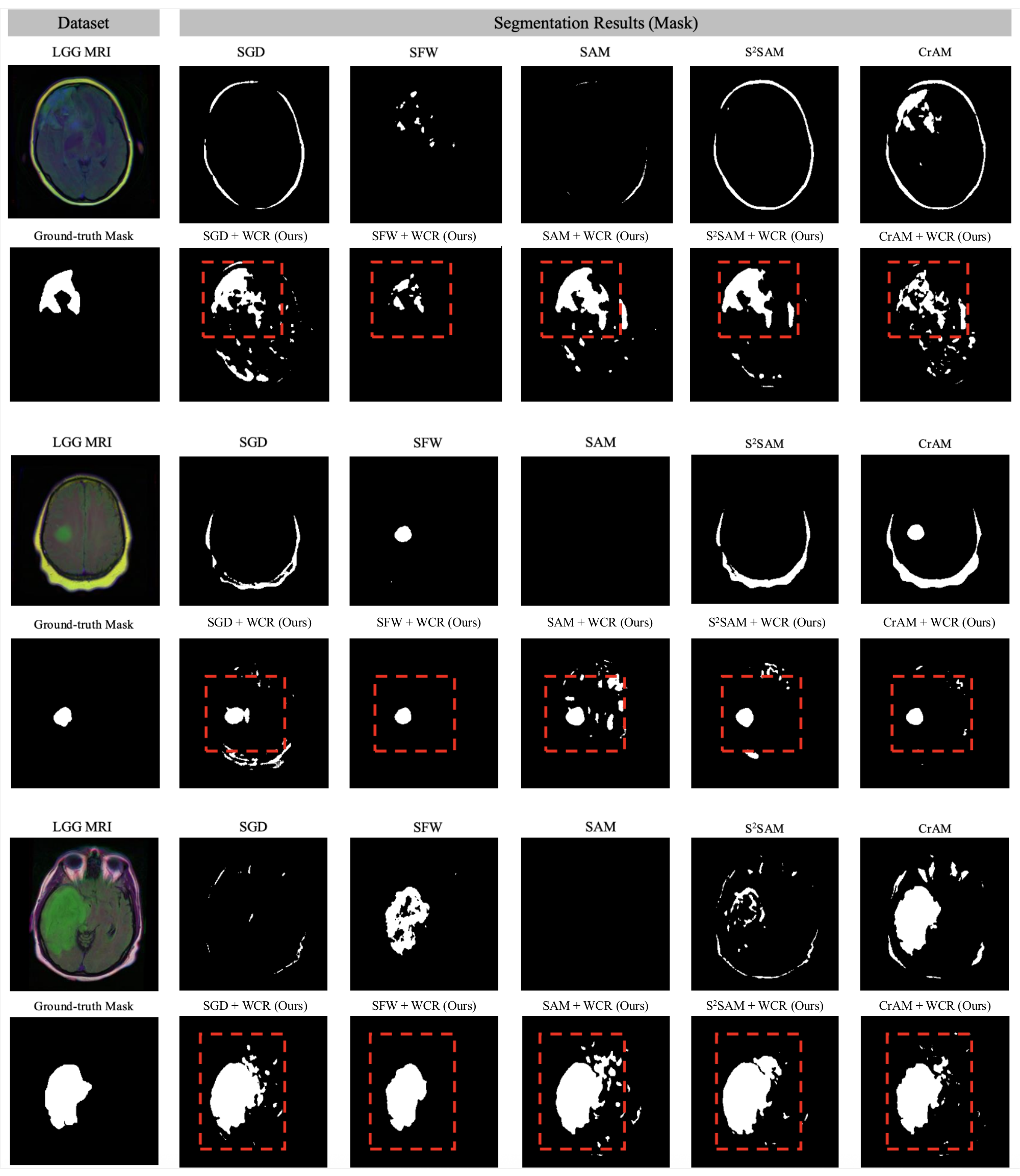}
  \caption{Qualitative segmentation results on the LGG MRI dataset using the ResNet-50–UNet architecture under 85\% pruning. Each column displays the outputs of different pruning-robust training methods, shown both with and without our proposed Weight Concentration Regularizer (WCR), along with the corresponding ground-truth mask.}
  \label{fig:append-tumor1}
\end{figure*}

\begin{figure*}[t]
  \centering
  \includegraphics[width=1\textwidth]{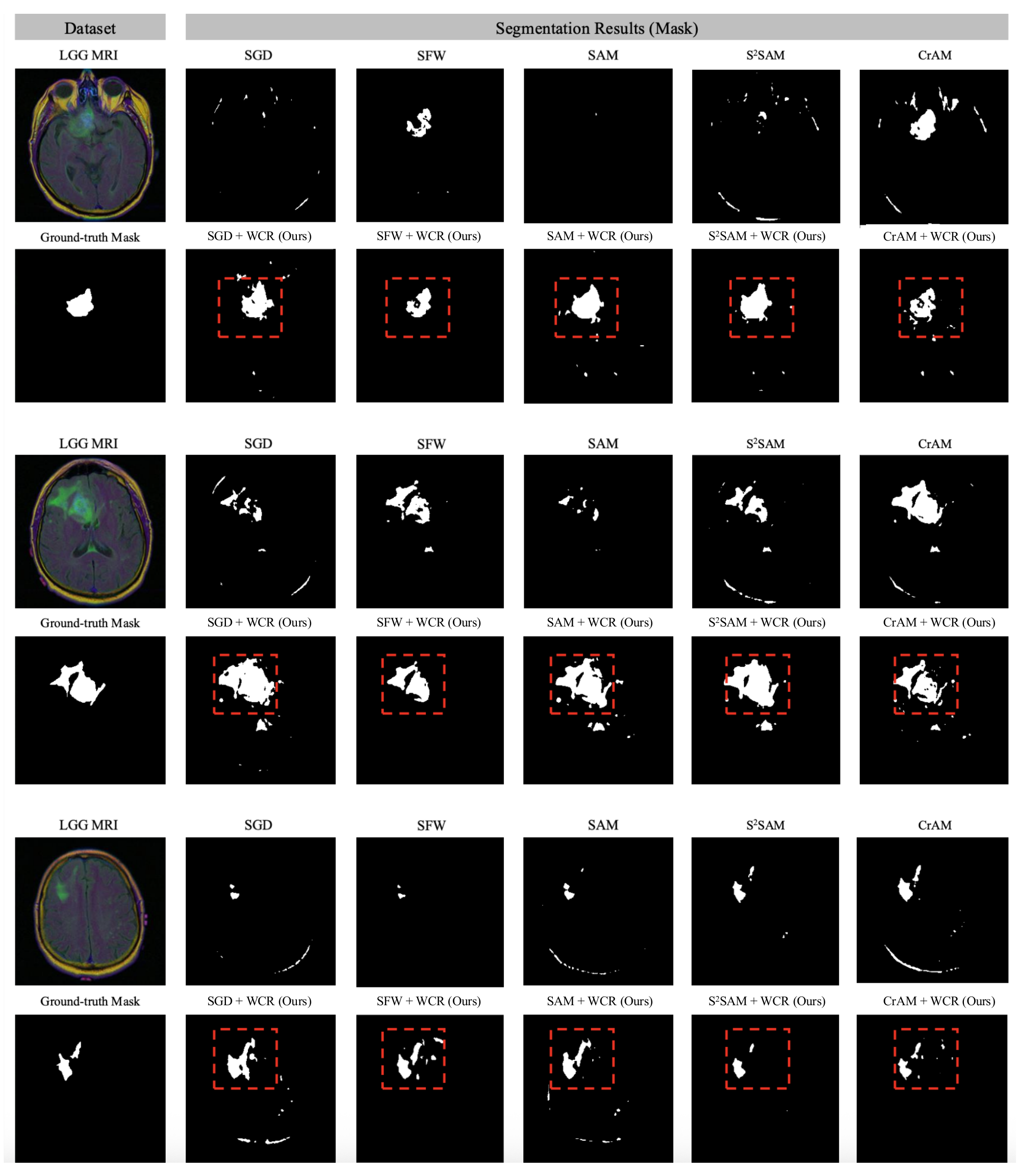}
  \caption{Qualitative segmentation results on the LGG MRI dataset using the ResNet-50–UNet architecture under 85\% pruning. Each column displays the outputs of different pruning-robust training methods, shown both with and without our proposed Weight Concentration Regularizer (WCR), along with the corresponding ground-truth mask.}
  \label{fig:append-tumor2}
\end{figure*}

\subsection{Additional segmentation results}
\label{appendix:seg}
To further examine the effect of the proposed Weight Concentration Regularizer (WCR) under high pruning rates, we provide additional qualitative segmentation results in Figure~\ref{fig:append-tumor1} and Figure~\ref{fig:append-tumor2}. Each figure compares the segmentation masks produced by several pruning-robust training methods, both with and without WCR, alongside the corresponding ground-truth tumor annotations.

Across all examples, models trained without WCR often show degraded predictions after 85\% one-shot pruning. Typical failure patterns include incomplete tumor regions, loss of boundary continuity, and irregular or noisy shapes. In contrast, models trained with WCR consistently preserve more complete lesion structures and maintain clearer and more accurate tumor boundaries. Methods such as SFW, SAM, and S$^2$SAM particularly benefit from the addition of WCR, which stabilizes the model's internal representations even under severe sparsity.

These qualitative findings are consistent with the quantitative results presented in the main paper. By promoting broader and more structured weight distributions during training, WCR reduces the distortion caused by aggressive pruning and leads to better segmentation masks, especially for tumors with complex shapes or low-contrast regions.

\end{document}